%% file: neurips_2026.tex
\documentclass{article}

\PassOptionsToPackage{numbers, compress}{natbib}

\usepackage[preprint]{neurips_2026}

\usepackage[utf8]{inputenc} 
\usepackage[T1]{fontenc}    
\usepackage{hyperref}       
\usepackage{url}            
\usepackage{booktabs}       
\usepackage{amsfonts}       
\usepackage{nicefrac}       
\usepackage{microtype}      
\usepackage{xcolor}         
\usepackage{algorithm}
\usepackage{algpseudocode}
\usepackage{amsmath}
\usepackage{graphicx}
\usepackage{subcaption}
\usepackage{enumitem}
\usepackage[table]{xcolor}
\usepackage[table]{xcolor}  
\usepackage{multirow}        
\usepackage{array}           
\usepackage{booktabs}        
\usepackage{float}     
\usepackage{placeins}  
\usepackage{caption}   
\definecolor{ourshl}{RGB}{230,242,230} 
\usepackage{tcolorbox}
\usepackage{amssymb}
\usepackage{pifont}

\newcommand{\cmark}{\ding{51}}
\newcommand{\xmark}{\ding{55}}

\newcounter{box}

\algrenewcommand{\algorithmiccomment}[1]{// #1}

\newcommand{\notrun}{\rule{1.5em}{0.4pt}\textsuperscript{\dag}}
\title{Auto-Dreamer: Learning Offline Memory Consolidation for Language Agents}

\author{%
  Chongrui Ye\thanks{Equal contribution. Order determined by coin flip; both authors reserve the right to list themselves as first author.}$^{,1}$\;\;
  Yuxiang Liu\footnotemark[1]$^{,1}$\;\;
  Yu Wang$^{2}$\;\;
  Haofei Yu$^{1}$ \\
  \textbf{Yining Zhao$^{1}$\;\;
  Ge Liu$^{1}$\;\;
  Julian McAuley$^{2}$\;\;
  Jiaxuan You$^{1}$} \\[2pt]
  $^{1}$University of Illinois Urbana-Champaign \quad
  $^{2}$University of California San Diego
}

\usepackage{listings}
\usepackage{xcolor}

\begin{document}

\maketitle

\begin{abstract}
Language agents increasingly operate over streams of related tasks, yet existing memory systems struggle to convert accumulated experience into reusable knowledge. Retrieval-augmented and structured memory methods record per-session observations effectively, but often couple acquisition and consolidation into a single online process, leaving the agent without a global view across sessions to discover recurring patterns, abstract shared procedures, or prune redundant entries. Inspired by complementary learning systems theory, we propose Auto-Dreamer, a learned offline consolidator for language-agent memory. Auto-Dreamer decouples fast per-session memory acquisition from slow cross-session consolidation. Given a selected working region of a typed memory bank, the consolidator treats the region as read-only evidence, performs bounded tool-use to inspect entries and provenance-linked source trajectories, and synthesizes a fresh compact replacement set that abstracts across sessions and supersedes the original region.
We train Auto-Dreamer via GRPO, using end-to-end agent performance as the reward signal to learn how to consolidate memories acquired through fast online experience. 
Trained on ScienceWorld trajectories alone, Auto-Dreamer outperforms fixed, RL-trained, and prompted memory baselines on ScienceWorld by 7 points while using an active memory bank 12$\times$ smaller than the strongest baseline, and continues to lead on held-out ALFWorld and WebArena without retraining --- using 6$\times$ less memory than the strongest baseline on ALFWorld.
\end{abstract}

\input{sections/1_introduction}
 
\input{sections/2_related_work}

\input{sections/3_preliminaries}
\input{sections/4_method}

\input{sections/5_experiments}

\input{sections/6_conclusion}
\bibliographystyle{plainnat}
\bibliography{references}

\newpage
\appendix
\input{appendix}

\clearpage
\end{document}

%% file: sections/1_introduction.tex
\section{Introduction}
\label{sec:intro}
Language agents are increasingly deployed over streams of related tasks rather than isolated interactions~\citep{llm_agent_survey,xi2023rise}. In such settings, long-term memory is not merely a retrieval cache for past entities or user preferences; it is the mechanism by which an agent converts raw experience into reusable procedures, environment knowledge, and behavioral priors that improve future decision making. A memory system must therefore solve two distinct problems: it must rapidly acquire useful information from each new trajectory, and it must periodically reorganize accumulated experience into a form that is compact, non-redundant, and useful for future tasks.

Recent work has made substantial progress on individual components of language-agent memory~\citep{hu2025memorysurvey,huang2026rethinking}, including retrieval-augmented episodic stores~\citep{zhong2024memorybank,letta}, structured memory systems~\citep{mirix,amem}, procedural skill libraries~\citep{memp,wang2023voyager,reasoningbank}, reflection-based methods~\citep{shinn2023reflexion,madaan2023self}, and RL-trained memory managers~\citep{umem,remem,memoryr1,memalpha}. Despite this progress, two challenges remain. First, a consolidation problem: existing methods typically couple acquisition and consolidation into a single online update process, so each update is made with limited evidence from the current session. This makes it difficult to discover recurring patterns, abstract reusable procedures that generalize across sessions, resolve contradictions, or prune redundant entries. Second, a memory-utility problem: RL-trained memory methods optimize online construction or retrieval rather than offline consolidation under an explicit downstream utility objective, so they do not directly learn which memories are load-bearing, which entries are redundant, or how to trade off success against memory compactness.

We take inspiration from complementary learning systems (CLS) theory 
of human memory, in which a fast hippocampal system encodes 
individual episodes and a slower neocortical system gradually 
extracts shared structure across episodes~\citep{mcclelland1995cls,
kumaran2016cls,mcclelland2020integration}. We adopt CLS not as a 
biological claim about language models, but as an operational design 
principle for separating fast acquisition from slow cross-session 
consolidation. We introduce \textbf{Auto-Dreamer}, a learned offline 
consolidator for language-agent memory.\footnote{Auto-Dreamer is 
distinct from the Dreamer family of world models~\citep{
hafner2019dreamer,hafner2023dreamerv3}; our method operates on 
memory entries and source trajectories, not the latent environment 
dynamics.} Auto-Dreamer is the slow-timescale counterpart to a 
fast per-session writer. Given a typed memory bank produced by 
the writer, it performs a multi-step tool-use rollout: searching 
memory, inspecting candidate entries, retrieving raw source 
trajectories for provenance, and synthesizing new entries that 
abstract across sessions. Its core operation is \emph{region rewriting}: 
the consolidator treats a selected working region as read-only evidence 
and synthesizes a fresh replacement set that supersedes the original 
region. This replacement semantics makes compactness structural rather than auxiliary: old entries do not persist by default, and information survives only if it is re-synthesized into the
replacement set. As a result, abstraction, deduplication, contradiction resolution, and omission-based forgetting become default behaviors. We train Auto-Dreamer with 
GRPO~\citep{shao2024grpo} using a composite reward that combines 
downstream task performance with a counterfactual utility term 
estimated by random memory masking, which penalizes redundant 
entries while rewarding load-bearing ones. The task agent and the 
per-session writer remain fixed throughout training, isolating 
the contribution of the consolidator.

We evaluate Auto-Dreamer in two regimes: continual-memory deployment, where the bank starts empty and grows over the task stream, and fixed-bank consolidation, where a pre-built bank is rewritten once. The results support three conclusions. First, Auto-Dreamer improves task success while maintaining substantially smaller active memory banks: in continual deployment, it achieves 41.1\% success on ScienceWorld~\citep{wang2022scienceworld}, 7 points above the strongest baseline with 12$\times$ less memory; 60.2\% on held-out ALFWorld~\citep{shridhar2021alfworld} with 6$\times$ less memory than the strongest baseline; and 52.3\% on held-out WebArena~\citep{zhou2024webarena}, leading all baselines. Second, the learned consolidator transfers beyond its training distribution: although trained only on ScienceWorld trajectories, it improves performance on held-out ALFWorld and WebArena without further updates, including across a writer-backbone shift from Qwen3-14B~\citep{qwen3technicalreport} to Gemini-3.1-flash-lite-preview~\citep{gemini}. Third, controlled fixed-bank experiments and ablations show that the gains come from offline consolidation itself: region rewriting improves the quality of a given memory bank, while the counterfactual utility term suppresses redundant memories without sacrificing task performance.

Our contributions are summarized as follows:
\begin{itemize}[nosep,topsep=0pt,leftmargin=*]
\item \textbf{A two-timescale formulation of language-agent memory.}
We distinguish fast per-session acquisition from slow cross-session
consolidation and formulate the latter as a learned decision problem
over accumulated evidence.
\item \textbf{Region rewriting as a compactness-inducing consolidation primitive.}
We formulate offline consolidation as provenance-grounded region rewriting: a selected working region is treated as read-only evidence and replaced by a synthesized replacement set. This
differs from per-entry CRUD by making cross-session abstraction, deduplication, and omission-based forgetting the default update semantics.
\item \textbf{RL training with region-local credit.}
Because region rewriting produces a self-contained replacement set, we can evaluate it directly and assign local credit from downstream task performance without supervised memory labels. We further use counterfactual masking to favor load-bearing memories and suppress redundant entries, improving the task utility of the compact bank.
\end{itemize}

%% file: sections/2_related_work.tex
\vspace{-2mm}
\section{Related works}
\label{sec:related}
\vspace{-2mm}

\textbf{Memory systems for language agents.}
A growing body of work designs memory architectures for language agents. Early systems organize memory around atomic units or flat stores, such as A-MEM~\citep{amem}, Mem0~\citep{mem0}, MemOS~\citep{memos}, and SimpleMem~\citep{simplemem}. More recent work introduces richer typed memory spanning episodic, semantic, and procedural stores, including EverMemOS~\citep{evermemos}, MIRIX~\citep{mirix}, Nemori~\citep{nemori}, and PlugMem~\citep{plugmem}. A complementary line focuses on extracting reusable procedures or strategies from trajectories: Memp~\citep{memp} and Voyager~\citep{wang2023voyager} build procedural skill libraries, ExpeL~\citep{expel} extracts cross-task insights from successful and failed trajectories, ReasoningBank~\citep{reasoningbank} distills high-level reasoning strategies, and ReMem~\citep{remem} studies test-time memory evolution. These systems improve how agents store experience, but their memory updates are governed by prompted heuristics applied within or immediately after each session, without explicit cross-session consolidation.

\textbf{RL-trained memory managers.}
Recent work explores training language models to construct memory using reinforcement learning. MEM1~\citep{mem1} and MemAgent~\citep{memagent} train models to update simple, text-only memories. Memory-R1~\citep{memoryr1}, Learn-to-Memorize~\citep{learn-to-memorize}, REMEMBER~\citep{remember}, and Mem-$\alpha$~\citep{memalpha} introduce richer memory representations and teach agents to manage complex memory systems through interaction and feedback. However, these methods primarily focus on teaching the model to extract and organize knowledge from its input, rather than on improving downstream agentic task performance. Later methods bridge this gap: UMEM~\citep{umem} jointly trains memory extraction and management with GRPO under an online single-step interface, and MemRL~\citep{memrl} trains the agent to retrieve the correct memory at decision time. Nevertheless, all of these operate online: memory updates are interleaved with task execution, so consolidation evidence is limited to the current session. Auto-Dreamer addresses a complementary problem, operating offline over a bank accumulated across many sessions with access to the full memory bank and raw source trajectories.

\textbf{Offline computation and sleep-time memory.}
Sleep-time compute~\citep{lin2025sleeptime} pre-computes over persistent context before queries arrive, amortizing reasoning across future interactions. LightMem~\citep{lightmem} combines an online writer with periodic offline consolidation, but implements consolidation as a fixed prompted pipeline with per-entry CRUD decisions. Auto-Dreamer instead performs \emph{region rewriting}: it treats a selected working region as read-only evidence, then uses a learned multi-step tool-using consolidator to synthesize a fresh compact replacement set that abstracts across sessions and supersedes the original region. The replacement set is grounded in re-readable source trajectories and trained with downstream task reward.

%% file: sections/3_preliminaries.tex
\vspace{-2mm}
\section{Preliminaries}
\vspace{-2mm}
\label{sec:prelim}

\begin{figure}[t]
    \centering
    \includegraphics[width=\linewidth]{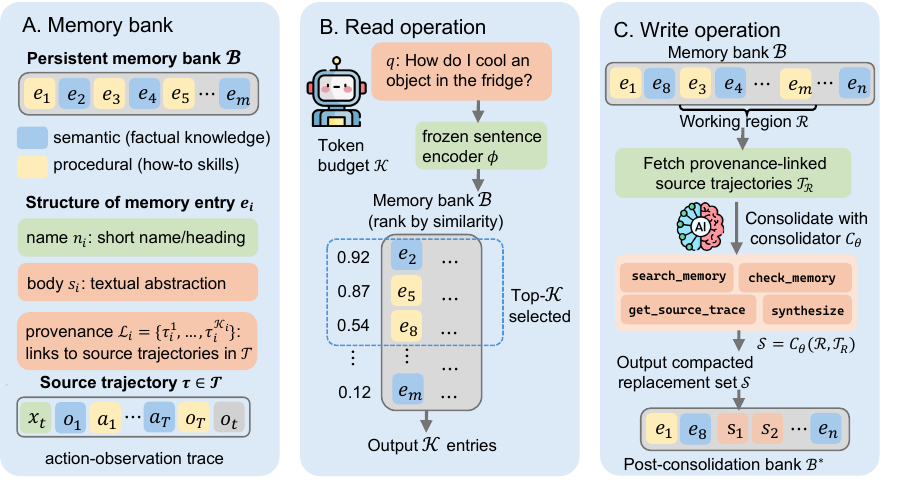}
    \caption{\textbf{Memory primitives and operations.}
    \textbf{(A)} The memory bank $\mathcal{B}$ holds typed entries (semantic 
    or procedural); each entry has a short name $n_i$, a body $s_i$, and 
    provenance links to source trajectories in the trajectory log $\mathcal{T}$.
    \textbf{(B)} The read operator retrieves the top-$K$ entries by cosine 
    similarity between a frozen sentence encoder $\phi$ applied to the query 
    and to each entry's name-body text.
    \textbf{(C)} The write operator applies a learnable consolidator $C_\theta$ 
    to a working region $\mathcal{R} \subseteq \mathcal{B}$ and its 
    provenance-linked trajectories $\mathcal{T}_{\mathcal{R}}$, producing a 
    replacement set $\mathcal{S}$ that supersedes $\mathcal{R}$ in the 
    post-consolidation bank $\mathcal{B}^{\star}$.}
    \label{fig:preliminaries}
\vspace{-3mm}
\end{figure}

\textbf{Task setup.}\label{sec:prelim_setup}
A frozen task agent operates over a stream of sessions $\tau$,
each yielding an action--observation trace and final outcome. The agent
has access to a typed long-term memory bank $\mathcal{B}$ through a fixed
retriever, and a trajectory log $\mathcal{T}$ that records raw sessions
for provenance. Offline consolidation leaves the task agent, retriever,
and memory schema fixed; it transforms $\mathcal{B}$ into a
post-consolidation bank $\mathcal{B}^{\star}$. Raw trajectories in
$\mathcal{T}$ are not retrieved by the task agent at decision time but
remain available to the consolidator as provenance evidence.

\textbf{Memory bank.}
A memory entry is a typed textual abstraction, with a short name $n$,
a body $s$, and provenance links to source trajectories in $\mathcal{T}$.
Each entry is either \emph{semantic} (factual environment knowledge,
e.g., \textit{``the toiletpaperhanger is typically on the bathroom
wall''}) or \emph{procedural} (reusable how-to skills, e.g.,
\textit{``to cool an object, place it in the fridge, wait, then
retrieve it''}). The memory bank $\mathcal{B}$ is a set of such entries.

\textbf{Memory operations.}
The bank supports two complementary operations: an online read operator
used by the frozen task agent, and an offline write operator learned by the
consolidator:
\begin{equation}
\label{eq:memory_ops}
\mathrm{Read}(q;\mathcal{B})
=
\mathrm{Top}\text{-}K_{e\in\mathcal{B}}
\cos\bigl(\phi(q),\phi(n_e\oplus s_e)\bigr), \;
\mathrm{Write}(\mathcal{B},\mathcal{R},\mathcal{T}_{\mathcal{R}})
=
(\mathcal{B}\setminus\mathcal{R})
\cup C_\theta(\mathcal{R},\mathcal{T}_{\mathcal{R}})
\end{equation}
Here $\phi$ is a frozen sentence encoder, $K$ is the largest ranked prefix
fitting the token budget, $\oplus$ denotes string 
concatenation, $\mathcal{R}\subseteq\mathcal{B}$ is the rewritten
memory region, and $\mathcal{T}_{\mathcal{R}}$ collects the source trajectories linked 
from entries in $\mathcal{R}$, accessible to the consolidator via provenance
with $C_\theta(\mathcal{R},\mathcal{T}_{\mathcal{R}})=\mathcal{S}$
producing a replacement set, yielding
$\mathcal{B}^{\star}=(\mathcal{B}\setminus\mathcal{R})\cup\mathcal{S}$.
We refer to this write operation as \emph{region rewriting}: entries in
$\mathcal{R}$ are treated as read-only evidence rather than edited in
place, and entries outside $\mathcal{R}$ are left unchanged.

\textbf{Evaluation tasks.}
At training time, the consolidator is scored on an evaluation task
set $\mathcal{V}$ randomly drawn from the training environment, disjoint from the
trajectories used to construct training memory regions and from all tasks
reported in Section~\ref{sec:experiments}.

%% file: sections/4_method.tex
\vspace{-2mm}
\section{Methodology}
\vspace{-2mm}
\label{sec:method}

\begin{figure*}[t]
\centering
\includegraphics[width=\linewidth]{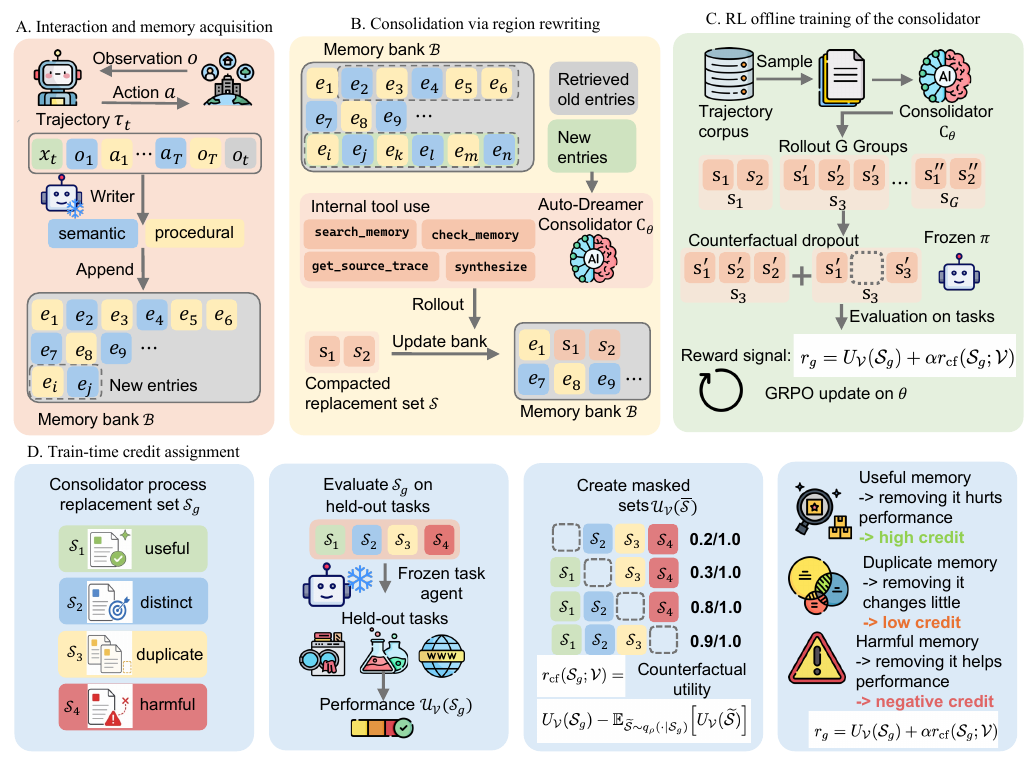}
\caption{\textbf{Auto-Dreamer overview.}
\textbf{(A)} A frozen writer appends typed entries from each trajectory 
$\tau_t$ to the memory bank $\mathcal{B}$.
\textbf{(B)} Every $k$ sessions, the consolidator $C_\theta$ rewrites a 
working region $\mathcal{R}$ into a replacement set $\mathcal{S}$ via 
tool-use rollout over memory and provenance trajectories.
\textbf{(C)} Training: $G$ group rollouts produce candidates 
$\{\mathcal{S}_g\}$, scored on evaluation tasks $\mathcal{V}$; GRPO 
updates $\theta$ using reward 
$r_g = R_{\mathcal{V}}(\mathcal{S}_g) + \alpha \, r_{\mathrm{cf}}(\mathcal{S}_g; \mathcal{V})$.
\textbf{(D)} The counterfactual term $r_{\mathrm{cf}}$ compares each 
$\mathcal{S}_g$ against masked variants, assigning high credit to 
useful entries, low credit to duplicates, and negative credit to 
harmful entries.}
\label{fig:method}
\vspace{-4mm}
\end{figure*}

\textsc{Auto-Dreamer} instantiates the offline consolidation problem
(Section~\ref{sec:prelim_setup}) with a learned tool-using policy.
A fixed online writer provides fast acquisition by extracting typed memories
from individual sessions. A learned offline consolidator provides slow
consolidation by rewriting selected memory regions after experience has
accumulated. The task agent, retriever, writer, memory schema, and token
budget are all fixed; only the consolidator parameters are updated during
training, and only the memory bank is updated after deployment. At a consolidation event, a region selector provides $\mathcal{R} \subseteq
\mathcal{B}$. The consolidator $C_\theta$ performs a bounded tool-use rollout:
searching the bank, inspecting candidate entries, and retrieving provenance-linked
source trajectories, and emitting synthesized entries. The synthesized entries form
a replacement set $\mathcal{S}$, inserted into the bank.

\vspace{-2mm}
\subsection{Designing Two-timescale Memory for Auto-Dreamer}
\vspace{-2mm}

\textbf{Fast online acquisition.}\label{sec:method_writer}
After each session, a prompted writer language model emits typed memory
entries that record potentially useful experience, each storing a
provenance pointer to the source trajectory. The writer is intentionally
local and append-only: it does not search the existing bank, compare
entries across sessions, or rewrite old memories. This favors
plasticity---new experience is recorded immediately and cheaply. Slower
cross-session operations (merging, abstraction, correction, compression,
forgetting) are delegated to the consolidator. The writer prompt and
schema are given in Appendix~\ref{sec:appendix-online-prompts}.

\textbf{Slow consolidation as region rewriting.}
\label{sec:method_dreamer}
A bank produced by repeated local writing is useful but inefficient. It may
contain duplicate procedures, overspecific rules, stale facts, and partial
observations whose common structure is only visible across sessions.
\textsc{Auto-Dreamer} addresses this by learning to rewrite active memory
regions. Given a region $\mathcal{R} \subseteq \mathcal{B}$ and its
provenance-linked trajectories $\mathcal{T}_{\mathcal{R}}$, the
consolidator performs a bounded tool-use rollout over a fixed turn
budget, with each step conditioned on $(\mathcal{R}, \mathcal{T}_{\mathcal{R}})$
and the history of previous tool calls and observations
(tool interface in Appendix~\ref{sec:appendix-dreamer}). The rollout
ends when the policy emits \texttt{terminate} or reaches the budget.
The synthesized entries form the replacement set $\mathcal{S}$, and the
bank is updated by replacing $\mathcal{R}$ with $\mathcal{S}$ according
to Eq.~\ref{eq:memory_ops}.

This provenance-grounded region-replacement semantics is central to Auto-Dreamer's compactness. In CRUD-style memory managers, existing memories persist unless the controller explicitly edits or deletes them, so consolidation is expressed as many local retention decisions. In region rewriting, the unit
of rewriting is a region rather than an individual entry: the old entries serve as evidence, and only information re-synthesized into the replacement set remains active. This makes abstraction, deduplication, contradiction resolution, and omission-based forgetting the default behaviors of the operator. As a result, compactness arises from the primitive itself, while learning determines which compact abstractions are most useful for downstream tasks.

\vspace{-2mm}
\subsection{Deploying Auto-Dreamer via Online Memory Acquisition}
\vspace{-2mm}
\label{sec:method_deployment}

In the online setting, the trained consolidator updates the memory bank
$\mathcal{B}$ using the Write operator from Eq.~\ref{eq:memory_ops}. We
trigger consolidation every $k$ sessions and define the working region
$\mathcal{R}$ as the union of entries newly written during the interval
and older entries retrieved by the task agent during the same interval.
This working region bridges online memory acquisition and offline
consolidation: newly written entries provide fresh evidence, while
recently retrieved entries identify older memories currently interacting
with the task agent's behavior. The consolidator treats $\mathcal{R}$ as
read-only evidence and synthesizes a replacement set $\mathcal{S}$,
yielding $\mathcal{B}^{\star}=(\mathcal{B}\setminus\mathcal{R})\cup\mathcal{S}$.
Entries outside $\mathcal{R}$ are left unchanged but may enter a future
working region if retrieved in subsequent tasks.


\vspace{-2mm}
\subsection{Training Auto-Dreamer via Offline Memory Consolidation}
\vspace{-2mm}
\label{sec:method_training}

We train the consolidator on regions sampled from an offline corpus of
agent trajectories. We first collect trajectories from the training
environments, run the fixed writer once, and store the resulting entries
together with their provenance links. At each training step, we sample
$J$ support trajectories $\{\tau^{(j)}\}_{j=1}^{J}$ and form the working
region $\mathcal{R}$ from their prewritten entries, with corresponding
source trajectories $\mathcal{T}_{\mathcal{R}}$ accessible through
provenance links. The consolidator samples $G$ tool-use rollouts over
$(\mathcal{R}, \mathcal{T}_{\mathcal{R}})$; rollout $g$ produces a
candidate replacement set $\mathcal{S}_g$. For direct credit assignment,
training evaluates each replacement set in a local bank consisting only
of the synthesized entries: $\mathcal{B}^{\star}_g = \mathcal{S}_g$.

\textbf{Local evaluation for credit assignment.}
Region rewriting turns consolidation into a locally evaluable operator-learning
problem. Each rollout produces a self-contained replacement set, so we evaluate it in a local bank consisting only of the
synthesized memories $ \mathcal{B}^{\star}_g = \mathcal{S}_g $.
This aligns the unit of
credit assignment with the unit produced by the policy: reward is assigned
directly to the replacement set, rather than to a full bank whose performance may be explained by memories not produced by the current rollout.
In this way, we learn a region-local improvement operator. At deployment, the same learned operator rewrites a selected working region, while entries outside the working region are left unchanged, as in
Eq.~\ref{eq:memory_ops}. The consolidation interface, provenance tools,
schema, frozen writer, retriever, task agent, and memory-token budget are
shared across training and deployment. Our continual-memory deployment experiments in
Section~\ref{sec:exp_online} show that this region-local objective transfers to persistent-bank composition.

We train the consolidator $C_\theta$ with GRPO~\citep{shao2024grpo}. The
writer, retriever, task agent, and memory-token budget are fixed across
rollouts; only the consolidator parameters $\theta$ are updated. The
consolidator is initialized from Qwen3-14B. Per-environment data
construction, rollout budgets, and optimization hyperparameters are given in
Appendix~\ref{app:training_hyperparameters}.

\textbf{Reward design.}
For rollout $g$, the reward combines downstream performance with a
counterfactual estimate of memory utility:
\begin{equation}
    r_g
    =
    U_{\mathcal{V}}(\mathcal{S}_g)
    +
    \alpha r_{\mathrm{cf}}(\mathcal{S}_g;\mathcal{V}).
    \label{eq:reward}
\end{equation}
The two terms are defined as
\begin{equation}
\label{eq:reward_terms}
\begin{array}{c@{\qquad}c}
\displaystyle
U_{\mathcal{V}}(\mathcal{S})
=
\frac{1}{|\mathcal{V}|}
\sum_{v\in\mathcal{V}}
\mathrm{Return}(v;\mathcal{S})
&
\displaystyle
r_{\mathrm{cf}}(\mathcal{S}_g;\mathcal{V})
=
U_{\mathcal{V}}(\mathcal{S}_g)
-
\mathbb{E}_{\widetilde{\mathcal{S}}\sim q_\rho(\cdot\mid\mathcal{S}_g)}
\!\left[
U_{\mathcal{V}}(\widetilde{\mathcal{S}})
\right].
\end{array}
\end{equation}
Here $\alpha$ weights the counterfactual term, and the distribution
$q_\rho(\widetilde{\mathcal{S}}\mid\mathcal{S}_g)$ masks a fixed fraction
$\rho$ of entries from $\mathcal{S}_g$ uniformly at random. In practice,
the expectation in $r_{\mathrm{cf}}$ is estimated with $M_g$ Monte Carlo
samples.
The counterfactual term measures the expected performance drop under random
ablation of the synthesized replacement set: masking load-bearing entries lowers
performance, masking redundant entries has little effect because duplicate
information remains, and masking harmful entries can improve performance,
making \(r_{\mathrm{cf}}\) negative. In the GRPO update, this favors replacement
sets whose entries improve downstream utility with minimal redundancy.

\begin{algorithm}[t]
\caption{Auto-Dreamer Consolidator Training}
\label{alg:training}
\setlength{\baselineskip}{0.92\baselineskip}
\begin{algorithmic}[1]
\Require Offline pool of trajectories with prewritten entries; frozen task agent; 
evaluation set $\mathcal{V}$; support size $J$; group size $G$; rollout budget $T_{\max}$
\For{each training step}
    \State Sample $J$ support trajectories $\{\tau^{(j)}\}_{j=1}^{J}$ from the offline pool
    \State Form working region $\mathcal{R} \leftarrow$ entries written from $\{\tau^{(j)}\}$, with provenance trajectories $\mathcal{T}_{\mathcal{R}}$
    \For{$g = 1, \ldots, G$}
        \State $\mathcal{S}_g \leftarrow C_\theta(\mathcal{R}, \mathcal{T}_{\mathcal{R}})$ \Comment{tool-use rollout, up to $T_{\max}$ steps}
        \State $r_g \leftarrow U_{\mathcal{V}}(\mathcal{S}_g) + \alpha \, r_{\mathrm{cf}}(\mathcal{S}_g; \mathcal{V})$ \Comment{Eq.~\ref{eq:reward}, local bank $\mathcal{B}^{\star}_g = \mathcal{S}_g$}
    \EndFor
    \State Update $\theta$ via GRPO using $\{r_g\}_{g=1}^{G}$
\EndFor
\end{algorithmic}
\end{algorithm}

%% file: sections/5_experiments.tex
\section{Experiments}
\label{sec:experiments}
\definecolor{hl-ours-a}{RGB}{235,243,250}
\definecolor{hl-ours-b}{RGB}{237,246,236}

We evaluate Auto-Dreamer along three axes. 
Section~\ref{sec:exp_online} studies realistic continual-memory deployment, showing improved task success over per-session memory baselines with a compact memory bank.
Section~\ref{sec:exp_offline} isolates the consolidation operator in a fixed-bank setting and shows gains over prompted offline baselines. 
Section~\ref{sec:exp_ablations} ablates the key design choices, highlighting the roles of offline consolidation and the counterfactual utility reward.

\subsection{Experimental Settings}
\label{sec:exp_setup}

\textbf{Tasks.}
We evaluate Auto-Dreamer on three tasks in different domains and difficulty: \textit{ALFWorld}~\citep{shridhar2021alfworld} (household instruction-following), \textit{ScienceWorld}~\citep{wang2022scienceworld} (text-based science experiments), and \textit{WebArena}~\citep{zhou2024webarena} (web navigation; shopping, shopping\_admin, gitlab).

\textbf{Models.}
The frozen task agent is shared across all methods within a domain: Qwen3.5-9B on ALFWorld and ScienceWorld, and Gemini-3-flash-preview~\citep{gemini3flash} on WebArena. For RL-trained memory baselines, we use the released 4B Mem-$\alpha$ checkpoint and a Qwen3-14B UMEM model trained on ALFWorld trajectories. All other baselines and Auto-Dreamer's per-session writer use Qwen3-14B on ALFWorld and ScienceWorld, and Gemini-3.1-flash-lite-preview~\citep{gemini} on WebArena, ensuring no baseline is disadvantaged by a weaker memory-generation LLM. The Auto-Dreamer consolidator $C_\theta$ is initialized from Qwen3-14B, trained on ScienceWorld trajectories only, and applied without further updates on all three domains---including across the writer-backbone shift on WebArena.

\textbf{Baselines.}
Ten baselines spanning seven families: no memory (No memory); reflective and insight extraction (Reflexion~\citep{shinn2023reflexion}, ExpeL~\citep{expel}); workflow and procedural memory (AWM~\citep{awm}, Memp~\citep{memp}); structured stores (ReasoningBank~\citep{reasoningbank}, Mem0~\citep{mem0}); two-timescale prompted memory (LightMem~\citep{lightmem}, the closest architectural counterpart to our method); RL-trained writers (Mem-$\alpha$~\citep{memalpha}, UMEM~\citep{umem}). Detailed descriptions and per-baseline implementation are in Appendix~\ref{app:baselines}.

\textbf{Metrics.}
We report task success rate (SR, \%), macro-averaged over task families within
each domain, and final active memory-bank size in tokens (\#Tok). For
continual-memory deployment, we additionally report the normalized area under
the cumulative success curve (AUC $\in [0,1]$) over the task stream.

\begin{table*}[t]
\centering
\small
\setlength{\tabcolsep}{2.1pt}
\renewcommand{\arraystretch}{1.04}
\caption{\textbf{Memory evaluation across continual deployment and controlled consolidation.}
Panel A reports continual-memory deployment where memory starts empty and is updated from trajectories collected during evaluation.
Panel B isolates the consolidation operator by giving each method the same fixed initial bank $\mathcal{B}_0$ and evaluating on held-out tasks with a frozen task agent.
For SR and AUC, higher is better; for \#Tok., lower is better.
\textbf{Bold} and \underline{underline} denote the best and second-best results.}
\label{tab:main_memory_eval}
\definecolor{hl-ours-a}{RGB}{235,243,250}
\definecolor{hl-ours-b}{RGB}{237,246,236}

\resizebox{\textwidth}{!}{%
\begin{tabular}{l ccc ccc ccc @{\hspace{16pt}}c@{\hspace{16pt}} l cc cc}
\toprule
\multicolumn{10}{c}{\textbf{(A) Continual-memory deployment}} &
&
\multicolumn{5}{c}{\textbf{(B) Bank consolidation: Control study}} \\
\cmidrule(lr){1-10} \cmidrule(lr){12-16}

\multirow{2}{*}{\textbf{Method}} &
\multicolumn{3}{c}{\textbf{ALFWorld}} &
\multicolumn{3}{c}{\textbf{ScienceWorld}} &
\multicolumn{3}{c}{\textbf{WebArena}} &
&
\multirow{2}{*}{\textbf{Method}} &
\multicolumn{2}{c}{\textbf{ALFWorld}} &
\multicolumn{2}{c}{\textbf{ScienceWorld}} \\

\cmidrule(lr){2-4} \cmidrule(lr){5-7} \cmidrule(lr){8-10}
\cmidrule(lr){13-14} \cmidrule(lr){15-16}

&
\textbf{SR} & \textbf{\#Tok.} & \textbf{AUC} &
\textbf{SR} & \textbf{\#Tok.} & \textbf{AUC} &
\textbf{SR} & \textbf{\#Tok.} & \textbf{AUC} &
&
&
\textbf{SR} & \textbf{\#Tok.} &
\textbf{SR} & \textbf{\#Tok.} \\
\midrule

\multicolumn{10}{l}{\cellcolor{gray!15}\textit{Baseline}} &
&
\multicolumn{5}{l}{\cellcolor{gray!15}\textit{Baseline}} \\
No memory & 30.8 & 0 & 0.287 & 28.7 & 0 & 0.295 & 44.6 & 0 & 0.527 &
&
No memory & 24.5 & 0 & 29.6 & 0 \\

\multicolumn{10}{l}{\cellcolor{gray!15}\textit{Reflective / insight extraction}} &
&
\multicolumn{5}{l}{\cellcolor{gray!15}\textit{Reflective / insight extraction}} \\
Reflexion~\citep{shinn2023reflexion} & 49.2 & 11,967 & 0.475 & 29.6 & 49,936 & 0.306 & 46.4 & 8,148 & 0.567 &
&
Reflexion~\citep{shinn2023reflexion} & 46.3 & \underline{455} & 31.0 & 661 \\
ExpeL~\citep{expel} & 33.6 & \textbf{2,042} & 0.306 & 28.3 & 11,628 & 0.289 & 50.8 & 3,371 & 0.576 &
&
ExpeL~\citep{expel} & \underline{70.1} & 2,942 & 38.1 & 808 \\

\multicolumn{10}{l}{\cellcolor{gray!15}\textit{Workflow / procedural memory}} &
&
\multicolumn{5}{l}{\cellcolor{gray!15}\textit{Workflow / procedural memory}} \\
AWM~\citep{awm} & 32.8 & 16,846 & 0.306 & 30.2 & \textbf{3,877} & 0.311 & \underline{52.0} & \textbf{890} & \underline{0.597} &
&
AWM~\citep{awm} & 66.9 & 758 & 33.6 & \textbf{103} \\
Memp~\citep{memp} & 31.4 & \underline{6,731} & 0.297 & 30.4 & 11,531 & 0.306 & 50.8 & 4,973 & 0.574 &
&
Memp~\citep{memp} & 67.3 & 927 & 40.7 & 351 \\

\multicolumn{10}{l}{\cellcolor{gray!15}\textit{Structured stores}} &
&
\multicolumn{5}{l}{\cellcolor{gray!15}\textit{Structured stores}} \\
ReasoningBank~\citep{reasoningbank} & 31.1 & 42,784 & 0.285 & 30.9 & 155,117 & 0.324 & 49.8 & 24,362 & 0.581 &
&
ReasoningBank~\citep{reasoningbank} & 43.8 & 3,021 & 25.0 & 3,221 \\
Mem0~\citep{mem0} & 30.6 & 119,013 & 0.279 & 26.8 & 89,854 & 0.281 & 50.3 & 43,355 & 0.571 &
&
Mem0~\citep{mem0} & 66.1 & 4,953 & \underline{42.0} & 1609 \\

\multicolumn{10}{l}{\cellcolor{gray!15}\textit{Two-timescale prompted}} &
&
\multicolumn{5}{l}{\cellcolor{gray!15}\textit{Two-timescale prompted}} \\
LightMem~\citep{lightmem} & 31.2 & 130,001 & 0.288 & 28.1 & 272,074 & 0.287 & \underline{52.0} & 370,874 & 0.567 &
&
LightMem~\citep{lightmem} & 40.6 & \textbf{243} & 31.8 & \underline{215} \\

\multicolumn{10}{l}{\cellcolor{gray!15}\textit{RL-trained writers}} &
&
\multicolumn{5}{l}{\cellcolor{gray!15}\textit{RL-trained writers}} \\
Mem-$\alpha$~\citep{memalpha} & 57.4 & 125,635 & \underline{0.566} & 30.0 & 344,599 & 0.297 & \multicolumn{3}{c}{\notrun} &
&
Mem-$\alpha$~\citep{memalpha} & 56.6 & 1,676 & 25.7 & 1,142 \\
UMEM~\citep{umem} & \underline{58.4} & 62,947 & 0.564 & \underline{34.1} & 80,918 & \underline{0.353} & \multicolumn{3}{c}{\notrun} &
&
UMEM~\citep{umem} & 54.2 & 5,184 & 29.3 & 1,298 \\

\multicolumn{10}{l}{\cellcolor{gray!15}\textit{Ours}} &
&
\multicolumn{5}{l}{\cellcolor{gray!15}\textit{Ours}} \\

\rowcolor{hl-ours-a}[\tabcolsep][\tabcolsep]
\textbf{\mbox{Auto-Dreamer}} &
\textbf{60.2} & 10,954 & \textbf{0.585} &
\textbf{41.1} & \underline{6,947} & \textbf{0.411} &
\textbf{52.3} & \underline{927} & \textbf{0.628} &
\cellcolor{white} &
\cellcolor{hl-ours-b}\textbf{\mbox{Auto-Dreamer}} &
\cellcolor{hl-ours-b}\textbf{72.7} & \cellcolor{hl-ours-b}634 &
\cellcolor{hl-ours-b}\textbf{44.3} & \cellcolor{hl-ours-b}539 \\
\bottomrule
\end{tabular}
}

\vspace{1mm}
\begin{minipage}{0.98\textwidth}
\scriptsize
\textsuperscript{\dag}UMEM and Mem-$\alpha$ rely on small open-source memory-optimizer models whose context windows cannot accommodate WebArena's accessibility-tree observations together with the memory bank. We do not retrain with larger backbones because doing so departs from the published configurations and exceeds our compute budget.
\end{minipage}

\end{table*}

\subsection{Main Results}
\label{sec:exp_online}

We evaluate Auto-Dreamer in the continual-memory deployment regime, where 
memory starts empty and is updated from trajectories collected during 
evaluation. After each completed session, the writer adds entries; methods 
with offline consolidation invoke their updater at a fixed cadence of $k$ sessions, with values reported in Appendix~\ref{app:evaluation_settings}. 

\textbf{Auto-Dreamer improves success while keeping memory compact.}
Table~\ref{tab:main_memory_eval} shows that Auto-Dreamer achieves the strongest
overall continual-memory performance across all three domains. On
ScienceWorld, it reaches 41.1\% SR, improving over the strongest baseline
UMEM by 7.0 points (34.1\%) and over the strongest prompted baseline
ReasoningBank by 10.2 points (30.9\%). Its AUC also improves from 0.353 to
0.411, indicating that the gain appears throughout the stream rather than
only at the end. The same pattern holds on ALFWorld, where Auto-Dreamer
achieves 60.2\% SR compared with 58.4\% for UMEM and 57.4\% for
Mem-$\alpha$, and on WebArena, where it obtains the highest SR despite the
longer-horizon tasks and noisier accessibility-tree observations.

\begin{figure*}[t]
\centering
\captionsetup[subfigure]{font=footnotesize,skip=1pt}
\begin{subfigure}[t]{0.30\textwidth}
    \centering
    \includegraphics[width=\linewidth]{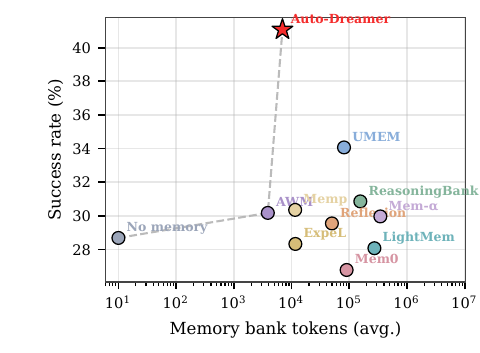}
    \caption{Success--cost tradeoff.}
    \label{fig:pareto}
\end{subfigure}
\hspace{0.018\textwidth}
\begin{subfigure}[t]{0.30\textwidth}
    \centering
    \includegraphics[width=\linewidth]{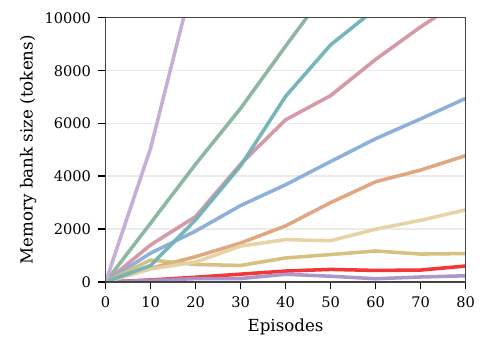}
    \caption{Bank growth; colors as in (a).}
    \label{fig:bank_growth}
\end{subfigure}
\hspace{0.018\textwidth}
\begin{subfigure}[t]{0.30\textwidth}
    \centering
    \includegraphics[width=\linewidth]{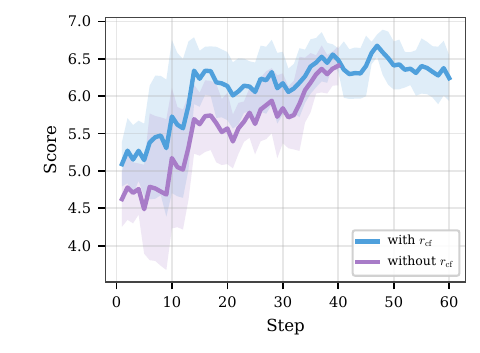}
    \caption{Dropout ablation: score.}
    \label{fig:dropout_score}
\end{subfigure}
\vspace{0.5mm}
\begin{subfigure}[t]{0.30\textwidth}
    \centering
    \includegraphics[width=\linewidth]{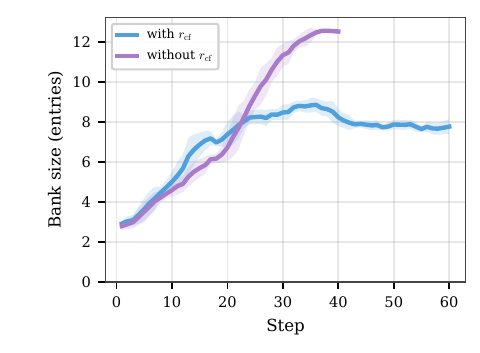}
    \caption{Dropout ablation: bank size.}
    \label{fig:dropout_bank}
\end{subfigure}
\hspace{0.018\textwidth}
\begin{subfigure}[t]{0.30\textwidth}
    \centering
    \includegraphics[width=\linewidth]{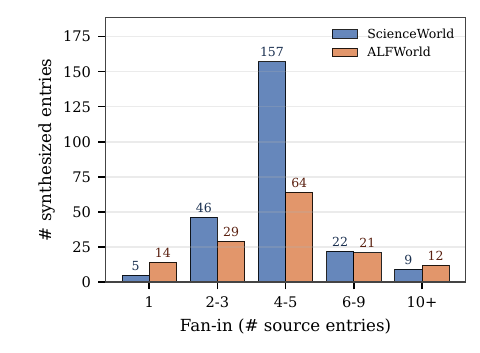}
    \caption{Provenance fan-in distribution.}
    \label{fig:fanin}
\end{subfigure}
\hspace{0.018\textwidth}
\begin{subfigure}[t]{0.30\textwidth}
    \centering
    \includegraphics[width=\linewidth]{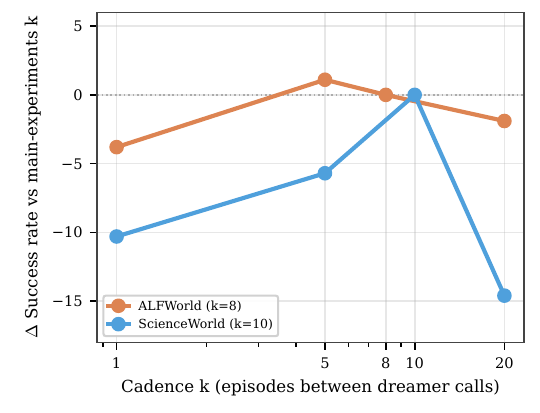}
    \caption{Consolidation cadence.}
    \label{fig:cadence_sweep}
\end{subfigure}
\vspace{-1mm}
\caption{
\textbf{Memory efficiency, reward ablation, and consolidator analysis.}
\textbf{(a)} Auto-Dreamer lies on the Pareto frontier of task success 
versus retrieval-time memory cost.
\textbf{(b)} Auto-Dreamer maintains a compact memory bank while most baseline 
methods grow monotonically as the task stream lengthens.
\textbf{(c,d)} The counterfactual utility reward preserves task 
performance while bounding bank growth during training.
\textbf{(e)} Provenance fan-in distribution peaks at fan-in $=5$ on both ScienceWorld and ALFWorld, indicating that synthesized entries typically draw on multiple source memories rather than being one-to-one copies.
\textbf{(f)} Consolidation cadence sweep: ScienceWorld performs best at the main cadence, while ALFWorld is comparatively robust across settings.
}
\label{fig:analysis}
\vspace{-2mm}
\end{figure*}

These gains are achieved with substantially lower retrieval-time memory cost.
On ScienceWorld, Auto-Dreamer uses 6.9k memory tokens, compared with 80.9k
for UMEM and 155.1k for ReasoningBank. On WebArena, it uses only 927 tokens,
compared with 370k for LightMem and 43.4k for Mem0, while still obtaining the
best SR and AUC. Figure~\ref{fig:pareto} summarizes this tradeoff: baselines
that approach Auto-Dreamer's success typically require much larger banks,
whereas compact baselines generally sacrifice success. Auto-Dreamer therefore
occupies the favorable region of the success--cost plane.

\textbf{The learned consolidator transfers across task domains and writer backbones.}
Auto-Dreamer's consolidator is trained only on ScienceWorld trajectories, yet
it improves continual-memory performance on held-out ALFWorld and WebArena
without additional updates. This demonstrates transfer not only across task
domains, but also across memory-acquisition backbones: the same trained
consolidator is paired with a Qwen3-14B writer on ALFWorld and ScienceWorld
and with a Gemini-3.1-flash-lite-preview writer on WebArena. This supports the
domain- and writer-agnostic design of the consolidation interface in
\S\ref{sec:method_dreamer}: the consolidator operates over typed textual
memory entries and provenance-linked trajectory excerpts, rather than
environment-specific state representations, action symbols, or writer-specific
hidden states.

The online results also test the training--deployment approximation in
\S\ref{sec:method_training}. Although training evaluates rewritten regions
locally for credit assignment, Panel~A evaluates repeated composition into a
persistent, growing bank. Auto-Dreamer's gains show that the locally trained
rewrite operator remains effective under retrieval competition and interaction
with older memories.


\subsection{Discussion on Bank Consolidation}
\label{sec:exp_offline}

The continual-memory deployment in \S\ref{sec:exp_online} measures end-to-end performance with many entangled factors: writer quality, retrieval competition over a growing bank, consolidation cadence, and downstream agent decisions over many sessions. We complement it with a controlled study that isolates the consolidation operator itself.

\textbf{Setup.}
Each method receives an identical fixed initial bank $\mathcal{B}_0$ constructed in advance from a sampled pool of trajectories run through the same fixed writer. The consolidator is invoked exactly once on $\mathcal{B}_0$, producing a consolidated bank $\mathcal{B}^\star$. We then evaluate the frozen task agent equipped with $\mathcal{B}^\star$ on a held-out task set drawn from the same environment family. Methods that lack a consolidation step (e.g., Reflexion, AWM) are evaluated directly on $\mathcal{B}_0$. This setup mirrors the training distribution of $C_\theta$ (\S\ref{sec:method_training}) and decouples consolidation quality from the dynamics of an evolving stream. We use ALFWorld and ScienceWorld; WebArena is omitted because the long-horizon, stateful nature of web tasks precludes constructing a meaningful fixed-bank evaluation that does not collapse into a stream.


\textbf{Auto-Dreamer also leads in the controlled regime.}
Panel B of Table~\ref{tab:main_memory_eval} reproduces the online ranking: Auto-Dreamer reaches 72.7\% on ALFWorld and 44.3\% on ScienceWorld, leading the strongest baseline on each (ExpeL at 70.1\%, Mem0 at 42.0\%) by 2.6 and 2.3 points respectively, with comparable or smaller memory cost. The narrower margin compared to continual deployment is consistent with the controlled regime removing the cumulative retrieval-competition effects that further amplify Auto-Dreamer's advantage at deployment time.




\subsection{Ablation Studies}
\label{sec:exp_ablations}

\textbf{Ablating offline consolidation.}
We ablate offline consolidation with two variants: \emph{writer-only}, 
which keeps the same task agent, retriever, schema, and per-session 
writer but disables consolidation; and \emph{untrained}, which uses 
the same region rewriting mechanism, tool-use rollout, and provenance grounding but 
without GRPO training.

Table~\ref{tab:ablation_consolidation} shows that the untrained 
pipeline already accounts for most of the bank-size reduction over 
writer-only (6--11$\times$ smaller across all three domains), 
supporting the view that region rewriting is the primary compactness 
mechanism: replacing a region with a synthesized set induces compression before learning. GRPO training then changes the quality and selectivity of the rewrite policy. 
On ScienceWorld and WebArena, training adds substantial SR gains 
(+9.7pp and +5.7pp); on ALFWorld, the aggregate gain is smaller 
(+1.0pp) and the active bank grows. Task family-level analysis 
reveals heterogeneous effects within ALFWorld: training helps 
on multi-step manipulation tasks but hurts on the single 
location-sensitive family \texttt{look\_at\_obj\_in\_light}, 
where task-specific location details are lost under abstraction. 
Excluding this family widens the trained-vs-untrained SR margin 
on the remaining five categories to $+4.9$pp, and we analyze 
this failure mode as Pattern~3 in \S\ref{sec:exp_qualitative}.

\begin{table}[h]
\centering
\small
\setlength{\tabcolsep}{3pt}
\renewcommand{\arraystretch}{0.95}
\caption{\textbf{Effect of offline consolidation}. Untrained Auto-Dreamer 
uses the same consolidator pipeline (region rewriting, tool-use 
rollout, provenance grounding) but without RL updates, comparing 
trained vs. untrained isolates the contribution of GRPO training.}
\label{tab:ablation_consolidation}
\begin{tabular}{l cc cc cc}
\toprule
\multirow{2}{*}{\textbf{Method}} &
\multicolumn{2}{c}{\textbf{ALFWorld}} &
\multicolumn{2}{c}{\textbf{ScienceWorld}} &
\multicolumn{2}{c}{\textbf{WebArena}} \\
\cmidrule(lr){2-3} \cmidrule(lr){4-5} \cmidrule(lr){6-7}
& \textbf{SR} & \textbf{\#Tok.}
& \textbf{SR} & \textbf{\#Tok.}
& \textbf{SR} & \textbf{\#Tok.} \\
\midrule
Writer-only & 48.0 & 50,735 & 29.6 & 61,679 & 44.2 & 6,421 \\
Auto-Dreamer (untrained) & 59.2 & 6,900 & 31.4 & 5,700 & 46.6 & 1,067 \\
\cellcolor{hl-ours-a}Auto-Dreamer
& \cellcolor{hl-ours-a}60.2
& \cellcolor{hl-ours-a}10,954
& \cellcolor{hl-ours-a}41.1
& \cellcolor{hl-ours-a}6,947
& \cellcolor{hl-ours-a}52.3
& \cellcolor{hl-ours-a}927 \\
\bottomrule
\end{tabular}
\end{table}

\textbf{Ablating counterfactual utility reward.}
We next train Auto-Dreamer with and without the counterfactual utility 
term $r_{\mathrm{cf}}$. Figures~\ref{fig:dropout_score} and 
\ref{fig:dropout_bank} report the resulting training dynamics. 
Without $r_{\mathrm{cf}}$, the consolidator continues to improve raw 
environment score, but the bank grows rapidly over training. 
With $r_{\mathrm{cf}}$, bank size first grows and then shrinks as 
training proceeds, while task performance remains competitive with 
the unshaped variant. This supports the intended role of counterfactual 
utility: it encourages compact, load-bearing memory banks 
without a measurable cost in task success.

\textbf{Ablating consolidation cadence.}
We sweep the consolidation cadence \(k \in \{1,5,20\}\) on ALFWorld and ScienceWorld, holding all other factors fixed. Figure~\ref{fig:cadence_sweep} reports \(\Delta\)SR relative to the main-experiment cadence (\(k=8\) on ALFWorld, \(k=10\) on ScienceWorld). ScienceWorld shows a larger cadence effect, with the main cadence outperforming both more frequent and less frequent consolidation. Performance is lower when consolidation is too frequent (\(k=1\)), suggesting that very small intervals provide insufficient cross-session evidence for useful abstraction. Performance also drops when consolidation is too infrequent (\(k=20\)), consistent with the consolidator being asked to process a larger and noisier working region that strains its context and tool-use budget. ALFWorld is more robust to cadence, with a range of roughly 5pp; \(k=5\) slightly outperforms the main setting by 1.1pp.

\subsection{Qualitative Analysis}
\label{sec:exp_qualitative}

\textbf{Multi-source consolidation.}
Figure~\ref{fig:fanin} reports the distribution of \emph{provenance fan-in}, the number of source entries cited by each synthesized output. On both ScienceWorld and ALFWorld, the distribution peaks at fan-in $=5$, indicating that synthesized entries typically draw on multiple source memories rather than being one-to-one copies. We next examine three qualitative patterns: two successful patterns drawn from matched tasks where the same agent succeeds with the Auto-Dreamer bank but fails with the writer-only bank, and one failure pattern where abstraction discards 
task-specific details.

\begin{tcolorbox}[colback=gray!5,colframe=gray!50,
                  title=\textbf{Case Study: Slot Abstraction (\texttt{find-entity}, task 90)},
                  fonttitle=\small,boxrule=0.4pt,arc=2pt,
                  left=4pt,right=4pt,top=4pt,bottom=4pt]
\small
\textbf{Task:} Find a living thing and place it in the \textbf{red box} in 
the \textbf{kitchen}. \\[2pt]
\begin{tabular}{p{0.46\linewidth} p{0.46\linewidth}}
\textbf{Writer memory (3 hardcoded entries)} & \textbf{Auto-Dreamer memory} \\[2pt]
\hangindent=1em\hangafter=1\textcolor{red!70!black}{\xmark} ``move to \textbf{blue} box in \textbf{living room}'' & 
\hangindent=1em\hangafter=1\textcolor{green!50!black}{\cmark} ``move to designated container (yellow, red, purple, or orange box)'' \\
\hangindent=1em\hangafter=1\textcolor{red!70!black}{\xmark} ``move to \textbf{green} box in \textbf{bathroom}'' & 
\hangindent=1em\hangafter=1\textcolor{green!50!black}{\cmark} ``teleport to target room (bedroom, workshop, etc.)'' \\
\hangindent=1em\hangafter=1\textcolor{red!70!black}{\xmark} ``move to \textbf{green} box in \textbf{bathroom}'' (duplicate) & 
\hangindent=1em\hangafter=1\textcolor{green!50!black}{\cmark} ``living things typically outside, greenhouse, or art studio'' \\[4pt]
\textbf{Task agent (fail, 8 steps):} & \textbf{Task agent (success, 6 steps):} \\[2pt]
\hangindent=1em\hangafter=1\textcolor{red!70!black}{\xmark} No memory matches (red, kitchen); agent searches workshop freezer (no living thing inside) & 
\hangindent=1em\hangafter=1\textcolor{green!50!black}{\cmark} Goes to greenhouse, finds cherry tree (a living thing) \\
\hangindent=1em\hangafter=1\textcolor{red!70!black}{\xmark} Wanders to kitchen, focuses on an \textbf{orange} (a fruit, not living) & 
\hangindent=1em\hangafter=1\textcolor{green!50!black}{\cmark} Carries cherry tree to kitchen, places in red box \\
\hangindent=1em\hangafter=1\textcolor{red!70!black}{\xmark} Score $-1.0$ & 
\hangindent=1em\hangafter=1\textcolor{green!50!black}{\cmark} Score $1.0$ \\
\end{tabular}
\end{tcolorbox}
\refstepcounter{box}\label{box:slot}

\textbf{Pattern 2: filtering wrong and contradicting entries via abstraction.}
Writer entries from different past tasks can encode mutually 
contradictory claims about the same task element, or contain phrasing 
errors carried over from an LLM-generated trace. The agent, treating 
writer memory as authoritative, can act on a wrong entry and become 
stuck. Rather than adjudicating among conflicting specifics or 
propagating phrasing errors, the consolidator drops these entries 
and emits a higher-level rule that preserves the shared task 
structure while leaving instance-specific answers to in-context 
reasoning.

\begin{tcolorbox}[colback=gray!5,colframe=gray!50,
                  title=\textbf{Case Study: Filtering Wrong Entries (\texttt{lifespan-compare}, task 61)},
                  fonttitle=\small,boxrule=0.4pt,arc=2pt,
                  left=4pt,right=4pt,top=4pt,bottom=4pt]
\small
\textbf{Task:} Find the animal with the \textbf{longest}, then the \textbf{shortest}, 
lifespan. \\[2pt]
\begin{tabular}{p{0.46\linewidth} p{0.46\linewidth}}
\textbf{Writer memory (contradictory + wrong)} & \textbf{Auto-Dreamer memory} \\[2pt]
\hangindent=1em\hangafter=1\textcolor{red!70!black}{\xmark} ``longest is \textbf{crocodile}'' & 
\hangindent=1em\hangafter=1\textcolor{green!50!black}{\cmark} ``compare lifespans of the listed candidates'' \\
\hangindent=1em\hangafter=1\textcolor{red!70!black}{\xmark} ``longest is \textbf{sea turtle}'' (contradicts above) & 
\hangindent=1em\hangafter=1\textcolor{green!50!black}{\cmark} ``focus on longest first, then shortest'' \\
\hangindent=1em\hangafter=1\textcolor{red!70!black}{\xmark} ``shortest is mouse'' & 
\hangindent=1em\hangafter=1\textcolor{green!50!black}{\cmark} ``focus the adult life stage'' \\
\hangindent=1em\hangafter=1\textcolor{red!70!black}{\xmark} ``focus on adult \textbf{and} baby elephant'' (writer error: trace only focused adult) & \\[4pt]
\textbf{Task agent (fail, 18 steps):} & \textbf{Task agent (success, 3 steps):} \\[2pt]
\hangindent=1em\hangafter=1\textcolor{red!70!black}{\xmark} Follows the elephant entry; outputs action \texttt{focus on adult adult elephant} & 
\hangindent=1em\hangafter=1\textcolor{green!50!black}{\cmark} Identifies elephant as the longest-lived candidate, focuses on it \\
\hangindent=1em\hangafter=1\textcolor{red!70!black}{\xmark} Env: ``no such entity''; agent retries 16 times, never adapts & 
\hangindent=1em\hangafter=1\textcolor{green!50!black}{\cmark} Identifies dragonfly as the shortest-lived, focuses on it \\
\hangindent=1em\hangafter=1\textcolor{red!70!black}{\xmark} Step cap reached, score $-1.0$ & 
\hangindent=1em\hangafter=1\textcolor{green!50!black}{\cmark} Score $1.0$ \\
\end{tabular}
\end{tcolorbox}
\refstepcounter{box}\label{box:contradict}

\textbf{Pattern 3 (failure mode): over-compression of locally useful details.}
We compare Auto-Dreamer with the untrained variant. Both perform 
region-rewriting, and they differ in whether the consolidator policy 
is trained. Although the trained consolidator improves performance in 
most task families, it underperforms the untrained consolidator on 
\texttt{look\_at\_obj\_in\_light}. In this task, concrete locations of 
target object and light source can help disambiguate where to search 
and where to examine the object. These details are episodic in form, 
but can still be useful for the current task. The trained consolidator 
tends to replace such task-specific paths with generic procedural 
entries, while the untrained consolidator retains more task-specific 
details. This suggests that the learned policy can over-compress 
specific facts that are locally useful. Quantitatively, this single 
category drives the bulk of the ALFWorld trained-vs-untrained gap in 
Table~\ref{tab:ablation_consolidation}: removing 
\texttt{look\_at\_obj\_in\_light} widens 
the SR margin on the remaining five families from $+1.0$pp to $+4.9$pp 
($65.8$ vs.\ $60.9$).

\begin{tcolorbox}[colback=gray!5,colframe=gray!50,
                  title=\textbf{Case Study: Over-Abstraction (\texttt{look\_at\_obj\_in\_light}, task 308)},
                  fonttitle=\small,boxrule=0.4pt,arc=2pt,
                  left=4pt,right=4pt,top=4pt,bottom=4pt]
\small
\textbf{Task:} Examine the alarmclock under the desklamp (alarmclock 
on desk-2, desklamp on desk-1). \\[2pt]
\begin{tabular}{p{0.46\linewidth} p{0.46\linewidth}}
\textbf{Auto-Dreamer (untrained) memory} & \textbf{Auto-Dreamer memory} \\[2pt]
\hangindent=1em\hangafter=1\textcolor{green!50!black}{\cmark} ``alarmclock on desk-2, desklamp on desk-1; take then go'' & 
\hangindent=1em\hangafter=1\textcolor{red!70!black}{\xmark} ``Key considerations for interacting with light sources'' \\
\hangindent=1em\hangafter=1\textcolor{green!50!black}{\cmark} ``avoid repeated examine of desk-1 when target is on desk-2'' & 
\hangindent=1em\hangafter=1\textcolor{red!70!black}{\xmark} ``Step-by-step process for examining an object'' \\[4pt]
\textbf{Task agent (success):} & \textbf{Task agent (fail):} \\[2pt]
\hangindent=1em\hangafter=1\textcolor{green!50!black}{\cmark} Takes alarmclock to desk-1, examines under desklamp & 
\hangindent=1em\hangafter=1\textcolor{red!70!black}{\xmark} Lacks location info; loops examining wrong desk \\
\hangindent=1em\hangafter=1\textcolor{green!50!black}{\cmark} score $1.0$ & 
\hangindent=1em\hangafter=1\textcolor{red!70!black}{\xmark} score $-1.0$ \\
\end{tabular}
\end{tcolorbox}
\refstepcounter{box}\label{box:overabstract}


%% file: sections/6_conclusion.tex
\section{Conclusion}
\label{sec:conclusion}

We presented \textbf{Auto-Dreamer}, a two-timescale memory system that pairs
fast per-session writing with learned offline region rewriting. By separating
online acquisition from slow cross-session consolidation, Auto-Dreamer matches
or exceeds the strongest memory baselines on ALFWorld, ScienceWorld, and WebArena
while maintaining substantially smaller active memory banks. A consolidator
trained only on ScienceWorld transfers to held-out domains and across a
writer-backbone shift, supporting the view that consolidation over textual memory
entries and source trajectories can be domain- and writer-agnostic.
Several extensions follow naturally. The current consolidator rewrites
one working region per event; future work could maintain longer-range bank
structure, jointly optimize retrieval and consolidation, or handle multimodal source
trajectories. More broadly, offline learned consolidation may be useful whenever
agent experience accumulates faster than it can be reorganized in-session.

%% file: appendix.tex
\section{Limitations}
\label{appendix:limitations}
\textbf{Evaluation scope.}
Our evaluation is restricted to three text-based agent environments 
sharing an LLM-mediated interface (ALFWorld, ScienceWorld, WebArena). 
We make no claims about transfer to settings with structured state 
representations, non-textual observations~\citep{chen-etal-2025-decisionflow}, or domains where memory 
must encode visual or multimodal evidence.

\textbf{Writer and schema dependence.}
The consolidator operates over entries written by a fixed prompted 
writer with a specific semantic/procedural schema. We hold these 
constant to isolate the consolidator, but robustness to alternative 
writers, schemas, or noisier provenance links remains untested. 
Information missed by the writer cannot generally be recovered by 
the consolidator unless source trajectories make it salient.

\textbf{Retrieval-budget sensitivity.}
Our main evaluation uses top-$K{=}3$ retrieval with a token cap on 
retrieved entries. This regime favors compact banks; methods that 
benefit from larger retrieval budgets may rank differently under 
looser constraints. Characterizing how the Pareto frontier shifts 
with the retrieval budget is left to future work.

\textbf{Surrogate training objective.}
The local-bank training objective (\S\ref{sec:method_training}) is 
a surrogate for deployment-time bank composition. While our online 
experiments show that the surrogate transfers, the formal 
relationship between local-bank ranking and full-bank ranking is 
not characterized.

\textbf{Variance.}
We report point estimates without seed or task-order variance. 
Several margins in Table~\ref{tab:main_memory_eval}, particularly 
on ALFWorld, are small enough that variance estimates would be 
useful for interpretation.

\FloatBarrier

\section{Experimental Details}
\label{appendix:experimental_details}

\subsection{Baseline Details}
\label{app:baselines}

We compare \textsc{Auto-Dreamer} against ten memory mechanisms for LLM agents,
spanning six families. For each baseline we describe what is written into memory,
how it is retrieved, and how it is consolidated. Unless otherwise noted, we use
the original authors' prompts and hyperparameters, and pair each method with the
same backbone task agent for fairness.

\paragraph{No memory.}
A memoryless baseline in which the task agent receives only the current
observation and the task instruction. It serves as a baseline that isolates
the contribution of any memory mechanism on top of the underlying policy.

\subsubsection{Reflective / Insight Extraction}

\paragraph{Reflexion~\citep{shinn2023reflexion}.}
After each trajectory, the agent generates a free-form natural-language
\emph{reflection} that diagnoses failures and proposes corrections. Reflections
are appended to a per-task buffer and prepended to the prompt on subsequent
attempts. There is no cross-task generalization or structured retrieval: memory
is task-local and grows monotonically until truncated by the context budget.

\paragraph{ExpeL~\citep{expel}.}
ExpeL distills successful and failed trajectories into a small set of
high-level \emph{insights} (rules of thumb) and a pool of in-context exemplars.
At inference time, the most relevant insights and exemplars are retrieved by
similarity to the current task and inserted into the prompt. Compared to
Reflexion, ExpeL transfers across tasks and emphasizes compact, generalizable
rules over per-episode reflections.

\subsubsection{Workflow / Procedural Memory}

\paragraph{AWM (Agent Workflow Memory)~\citep{awm}.}
AWM induces reusable \emph{workflows} -- abstracted action templates extracted
from successful trajectories -- and stores them in a workflow library. On a new
task, the most relevant workflows are retrieved and injected as procedural
guidance for the agent. Memory growth is tied to the diversity of induced
workflows rather than to the number of trajectories, leading to compact stores.

\paragraph{Memp~\citep{memp}.}
Memp builds a procedural memory by summarizing trajectories into stepwise
\emph{procedures} and indexing them for retrieval. It supports both addition
and revision of procedures as new evidence accumulates, occupying a middle
ground between purely episodic stores (Reflexion) and abstract workflow
libraries (AWM).

\subsubsection{Structured Stores}

\paragraph{ReasoningBank~\citep{reasoningbank}.}
ReasoningBank maintains a structured bank of \emph{reasoning traces} extracted
from prior episodes, organized to support semantic retrieval. Each entry
captures the chain-of-thought and key decision points of a trajectory, which
are surfaced to the agent on related future tasks. The store grows quickly
with experience, trading retrieval coverage for substantial token overhead.

\paragraph{Mem0~\citep{mem0}.}
Mem0 is a general-purpose long-term memory layer that extracts atomic
\emph{facts} and \emph{preferences} from interactions and stores them in a
queryable memory graph. Retrieval combines vector similarity with light
structural reasoning over the graph. We adapt Mem0 to the agentic setting by
treating each trajectory as an interaction stream from which memories are
distilled.

\subsubsection{Two-Timescale Prompted}

\paragraph{LightMem~\citep{lightmem}.}
LightMem separates memory operations into two timescales: a fast
\emph{working memory} that buffers recent context, and a slower
\emph{consolidation} step that periodically distills the buffer into long-term
notes. Both stages are fully prompted, with no learned components. This yields
a clean ablation point for two-timescale designs that does not rely on
reinforcement learning.

\subsubsection{RL-Trained Writers}

\paragraph{Mem-$\alpha$~\citep{memalpha}.}
Mem-$\alpha$ trains the \emph{memory writer} with reinforcement learning,
optimizing what to write so that downstream task success is maximized. The
reader/retrieval pipeline is held fixed, isolating the contribution of a
learned write policy. We omit Mem-$\alpha$ on WebArena: the released checkpoint relies on a small 
4B memory-optimizer backbone whose context window cannot accommodate WebArena's accessibility-tree observations together with the memory bank (see footnote in Table~\ref{tab:main_memory_eval}).

\paragraph{UMEM~\citep{umem}.}
UMEM (Unified Memory) similarly trains the writer with RL but unifies
episodic, procedural, and semantic memory into a single store with a learned
update operator. It represents the strongest RL-trained baseline in our
comparison. As with Mem-$\alpha$, we omit UMEM on WebArena due to context
window constraints.

\subsubsection{Baseline Implementation}
\label{app:baselines-impl}

All baselines share the Qwen2.5-3B last-token-hidden embedder
($2048$-d) and FAISS \texttt{IndexFlatIP} retrieval over L2-normalized
vectors, exposing memory through a common \texttt{read()/write()/reflect()}
interface. Table~\ref{tab:baseline-mods} lists what we changed and what
we preserved verbatim from each method's released code; commit hashes
are recorded in the top-of-file docstring of every
\texttt{baselines/*/memory.py}. Hyperparameters are gathered in
Table~\ref{tab:baseline-hparams}, and verbatim memory-construction
prompts in Appendix~\ref{app:baseline-prompts}.

\begin{table}[!htbp]
\centering\small
\setlength{\tabcolsep}{4pt}
\renewcommand{\arraystretch}{1.10}
\caption{\textbf{Per-baseline modifications relative to each method's released code.}}
\label{tab:baseline-mods}
\begin{tabular}{p{2.0cm} p{5.4cm} p{5.7cm}}
\toprule[1.1pt]
\textbf{Baseline} & \textbf{Adapted} & \textbf{Preserved verbatim} \\
\midrule
ReasoningBank
 & Encoder swapped to Qwen2.5-3B; LLM-as-Judge success label replaced
   by the env's gold reward; MaTTS scaling not implemented.
 & Successful / failed extraction prompts and the memory-injection
   banner (Tabs.~\ref{tab:prompt-rb-success}--\ref{tab:prompt-rb-banner}),
   from commit \texttt{ea65efd2}. \\
\midrule
ExpeL
 & Encoder swapped to Qwen2.5-3B; few-shot count raised from $2$ to $10$.
 & Compare-critique, all-success, and rule-operation templates
   (Tabs.~\ref{tab:prompt-expel-compare}--\ref{tab:prompt-expel-rules}). \\
\midrule
LightMem
 & Encoder swapped to Qwen2.5-3B; topic-segmentation (paper §3.2) not
   replicated.
 & Three-stage sensory$\to$STM$\to$LTM pipeline with LLMLingua-2
   compression; metadata-extraction and consolidation prompts
   (Tabs.~\ref{tab:prompt-lightmem-extract}--\ref{tab:prompt-lightmem-update}). \\
\midrule
Mem0
 & Encoder swapped to Qwen2.5-3B; \texttt{mem0.Memory()} bypassed
   (upstream drifted to a V3 additive pipeline); driven directly with
   the legacy ADD/UPDATE/DELETE/NONE prompts from commit
   \texttt{3fbc1c9a}.
 & Fact-retrieval and memory-update operator prompts
   (Tabs.~\ref{tab:prompt-mem0-extract}--\ref{tab:prompt-mem0-update}). \\
\midrule
AWM
 & Encoder unused at \texttt{read()} (full per-task-type workflow
   library prepended verbatim, no top-$k$).
 & Induction prompt (Tab.~\ref{tab:prompt-awm-induction}) from commit
   \texttt{8c0ff8cd}; per-task-type pools ($6$ ALFWorld families);
   harness-native \texttt{think:}/\texttt{act:} formatting. \\
\midrule
Memp
 & Encoder swapped to Qwen2.5-3B; cold-start build optional (defaults
   empty, grown via online \texttt{reflect()}); cosine $\geq 0.75$
   threshold (cosine-equivalent of upstream squared-L2 $\leq 0.5$).
 & Workflow-build and failure-Adjustment prompts
   (Tabs.~\ref{tab:prompt-memp-build}--\ref{tab:prompt-memp-adjust})
   from commit \texttt{3066a1b2}. \\
\midrule
Reflexion
 & Adapted to a \emph{cross-task-type} reflection-transfer regime
   (failure reflections keyed by \texttt{task\_type}, surfaced for new
   games of the same type), since our harness runs each game once
   rather than across multiple trials.
 & ReAct + reflection layering; reflector prompt
   (Tab.~\ref{tab:prompt-reflexion-reflector}); two canonical ReAct
   exemplars layered alongside reflections. \\
\bottomrule[1.1pt]
\end{tabular}
\end{table}

\begin{table}[!htbp]
\centering\small
\setlength{\tabcolsep}{4pt}
\caption{\textbf{Retrieval and generation hyperparameters per baseline.}
``Budget'' is the per-step LLM-call count for write/read/reflect.
``$T$'' is the sampling temperature.}
\label{tab:baseline-hparams}
\begin{tabular}{l p{8.6cm}}
\toprule[1.1pt]
\textbf{Baseline} & \textbf{Settings} \\
\midrule
ReasoningBank
 & Budget $0/0/2$;
   $T_{\text{action}}{=}0.0$, $T_{\text{rationale}}{=}0.1$,
   $T_{\text{extract}}{=}1.0$. \\
ExpeL
 & \texttt{COMPARE\_PAIRS\_CAP}${=}4$,
   \texttt{SUCCESS\_CRITIQUE\_NUM}${=}8$,
   \texttt{MAX\_NUM\_RULES}${=}10$,
   \texttt{NUM\_FEWSHOTS}${=}10$;
   $T_{\text{insight}}{=}T_{\text{reflect}}{=}0.0$. \\
LightMem
 & LLMLingua-2 compression $r{=}0.7$;
   \texttt{stm\_token\_threshold}${=}512$;
   sensory Jaccard window $=3$ at threshold $0.7$;
   \texttt{consolidation\_every\_n\_episodes}${=}10$. \\
Mem0
 & Budget $2/0/1$; both LLM calls at $T{=}0$. \\
AWM
 & Up to $5$ most-recent successes per task type;
   $T_{\text{induce}}{=}0.1$, \texttt{INDUCTION\_MAX\_TOKENS}${=}2048$. \\
Memp
 & Top-$K{=}10$; cosine threshold $0.75$;
   deprecation rule \texttt{hit}${\geq}3$ \textsc{and}
   \texttt{success/hit}${<}0.5$;
   $T_{\text{build}}{=}T_{\text{reflect}}{=}0.1$. \\
Reflexion
 & Reflect on failure only ($1$ call); $0$ on success;
   \texttt{max\_reflections\_per\_type}${=}5$. \\
\bottomrule[1.1pt]
\end{tabular}
\end{table}

\subsection{Evaluation Settings}
\label{app:evaluation_settings}  

  \paragraph{Evaluation protocol.}                                                                                                                                    
  We adopt a \emph{prequential (online streaming)} protocol implemented in
  \texttt{online\_memory.eval.run\_online}: held-out tasks are presented to                                                                                           
  the agent one at a time in a fixed seeded order; before task $t$, the                                                                                               
  agent retrieves from the bank built from tasks $1,\dots,t-1$, and after                                                                                             
  the task its trajectory is fed to the writer (and, on cadence, to the                                                                                               
  dreamer) so that future tasks see whatever new entries this trajectory                                                                                              
  produced. No task is ever replayed.                                                                                                                                 
                                                                                                                                                                      
\paragraph{Consolidation cadence.}
Methods with offline consolidation invoke their updater at a fixed cadence of
$k$ completed sessions. We use $k{=}10$ for ScienceWorld, $k{=}8$ for
ALFWorld, and $k{=}5$ for WebArena. The same cadence is used for all
offline-consolidation methods within each environment.

\paragraph{Environments and task pools.}
We evaluate on three long-horizon agent environments through a single
\texttt{EnvAdapter} interface so that the agent loop, retrieval pipeline,
and memory store are byte-identical across domains
(Table~\ref{tab:eval_envs}). Tasks are deterministically seeded with
\texttt{-{}-seed 42}; ALFWorld and ScienceWorld additionally set
\texttt{-{}-shuffle-tasks} to break the default
\texttt{task\_type$\to$variation} clustering and load a frozen episode list
via \texttt{-{}-episodes-path episodes.jsonl}.

\begin{table}[!htbp]
\centering
\small
\caption{Task pools used in online evaluation. ALFWorld and ScienceWorld
sub-pools are produced by sampling on
\texttt{task\_type}; WebArena combines the shopping, shopping\_admin,
and gitlab task families.}
\label{tab:eval_envs}
\begin{tabular}{l p{6cm} c c}
\toprule
\textbf{Env} & \textbf{Task pool} & \textbf{Size} & \textbf{\texttt{-{}-max-agent-steps}} \\
\midrule
ALFWorld     & \texttt{valid\_seen} + \texttt{valid\_unseen}, 6 task families & $274$ & $50$ \\
ScienceWorld & HF \texttt{test} split, 7 task types sampled (see Appx.~\ref{appendix:data-stats}) & $660$ & $100$ \\
WebArena     & shopping + shopping\_admin + gitlab task families & $117$ & $300$ \\
\bottomrule
\end{tabular}
\end{table}

\paragraph{Models served.}
Open-weight roles (task agent, writer when Qwen, dreamer, embedder) are 
served as independent SGLang endpoints on the same evaluation node 
(see Hardware below); the Gemini-3-flash-preview task agent and 
Gemini-3.1-flash-lite-preview writer used on WebArena are accessed via 
the Google AI API. All endpoints expose the OpenAI-compatible 
\texttt{/v1/chat/completions} interface. 
Decoding uses temperature $0.7$ and top-$p=0.9$, with
$\texttt{enable\_thinking}=\text{False}$ on Qwen3 endpoints.

\paragraph{Hardware.}
Training uses 8 NVIDIA H100 (80\,GB) GPUs. Evaluation runs use one node 
with 4 NVIDIA GH200 (96\,GB) GPUs serving all model endpoints (task 
agent, writer, dreamer, embedder).
                                                                                                                                                                      
  \paragraph{Memory store and retrieval.}
  Memory is persisted in LanceDB partitioned by \texttt{run\_id}, with                                                                                                
  $2048$-dim Qwen2.5-3B embeddings. At task start we issue a single
  retrieval against the task instruction and return up to top-$k=3$                                                                                                   
  entries (hybrid kNN $+$ salience rerank, $1500$-token budget); during
  the episode we additionally refresh memory every $8$ environment steps                                                                                              
  by re-querying with $\text{instruction}\,\Vert\,\text{last-K
  actions}\,\Vert\,\text{current observation}$ and appending the top-$1$                                                                                              
  retrieved entry to the user message. Retrieved entries are injected as                                                                                              
  raw \texttt{INSERT\_*} blocks inside a                                                                                                                              
  \texttt{===\,Memory from past experience\,===\,\ldots\,===\,End Memory\,===}                                                                                        
  panel appended to the env-native system prompt---the same format used at                                                                                            
  training time, so no method gains from prompt-format mismatch.                                                                                                      
                                                                                                                                                                      
  \paragraph{Logging and metrics.}
  For every task we record success, final environment score, episode                                                                                                  
  length, retrieved entry IDs and token count, writer/dreamer events, and
  bank size, streamed to \texttt{per\_task.jsonl},
  \texttt{trajectories.jsonl}, and \texttt{dreamer\_calls.jsonl}; the                                                                                                 
  aggregate \texttt{summary.json} reports success rate, mean final score,                                                                                             
  end-of-run active and retired bank sizes, total wall-clock time, and                                                                                                
  per-role LLM call and token counts. Unless noted, we report success rate                                                                                            
  and mean final score over the full task stream and plot bank size and                                                                                               
  dreamer firings indexed by task position $t$ to expose prequential
  learning dynamics rather than only end-of-stream aggregates.

\FloatBarrier

\subsection{Training Hyperparameters}
\label{app:training_hyperparameters}

Table~\ref{tab:training-hparams} reports the full set of training hyperparameters
for the ScienceWorld GRPO run, including model and optimization settings, rollout
and generation parameters, environment and episode configuration, and reward
shaping coefficients. The trained consolidator is applied to all three evaluation
domains without further updates.

\begin{table}[H]
\centering
\small
\caption{Training hyperparameters for the ScienceWorld GRPO run.}
\label{tab:training-hparams}
\begin{tabular}{ll}
\toprule
\textbf{Parameter} & \textbf{Value} \\
\midrule
\multicolumn{2}{l}{\textit{Model \& optimization}} \\
Base model                          & Qwen3-14B \\
Algorithm                           & GRPO (agent loop) \\
Optimizer                           & AdamW \\
Learning rate                       & $1\times 10^{-6}$ \\
LR schedule                         & constant \\
Warmup ratio                        & 0.0 \\
Entropy coefficient                 & 0.0 \\
KL loss coefficient ($\beta$)       & $1\times 10^{-3}$ \\
KL loss type                        & low-variance KL \\
Random seed                         & 42 \\
\#GPUs                              & 8 \\
Tensor-parallel size                & 1 \\
\midrule
\multicolumn{2}{l}{\textit{Rollout \& generation}} \\
Training steps                      & 200 \\
Batch size (prompts)                & 16 \\
Rollout group size $N$              & 8 \\
Sampling temperature (train)        & 1.0 \\
Top-$p$ (train)                     & 0.9 \\
Sampling temperature (val)          & 0.0 \\
Top-$p$ (val)                       & 1.0 \\
Max new tokens / turn               & 2{,}048 \\
Max prompt tokens                   & 3{,}072 \\
Max response tokens                 & 18{,}432 \\
Max model length                    & 21{,}504 \\
GPU memory utilization              & 0.80 \\
Thinking mode                       & disabled \\
\midrule
\multicolumn{2}{l}{\textit{Environment \& episode}} \\
Environment                         & ScienceWorld \\
Simplifications                     & teleport, openDoors, self-watering pots \\
Max env. steps / episode            & 50 \\
Max assistant / user turns          & 40 / 40 \\
History window (turns)              & 6 \\
Valid-action sample size            & 20 \\
Groups per prompt ($k$)             & 10 \\
Evaluations per sample              & 12 \\
Min. groups per task type           & 20 \\
\midrule
\multicolumn{2}{l}{\textit{Reward shaping}} \\
Format-penalty weight               & 0.5 \\
MC-dropout weight                   & 0.5 \\
MC-dropout: min $N$                 & 1 \\
MC-dropout: sample fraction         & 0.25 \\
MC-dropout: drop fraction           & 0.5 \\
Task model                          & Qwen3.5-9B \\
\bottomrule
\end{tabular}
\end{table}

\FloatBarrier

\section{Artifact Details}
\label{appendix:artifact_details}

\subsection{Model License}

\textbf{Gemini-3.1-flash-lite-preview} License: Proprietary\\
\textbf{Gemini-3-flash-preview} License: Proprietary\\
\textbf{Qwen3-14B} License: Apache 2.0\\
\textbf{Qwen3.5-9B} License: Apache 2.0

\subsection{Software Versions}
For our agentic RL environments, we build upon OpenTinker\footnote{\url{https://github.com/open-tinker/OpenTinker}}, which provides unified task configurations across ALFWorld and ScienceWorld. Our training infrastructure is built on top of verl\footnote{\url{https://github.com/volcengine/verl}}.

\subsection{Dataset Statistics}
\label{appendix:data-stats}
We report the dataset statistics for the three interactive benchmark 
environments used in our experiments: ALFWorld, ScienceWorld, and 
WebArena. ScienceWorld is used for both training-data 
construction and held-out evaluation; ALFWorld and WebArena are held-out only and not 
used during training. All environments' statistics follow the 
data-processing and environment configurations of the original papers. Table~\ref{tab:dataset_stats} summarizes the train and test 
splits; Table~\ref{tab:eval_envs} 
reports the evaluation pool.


\begin{table}[!htbp]
\centering
\small
\setlength{\tabcolsep}{6pt}
\renewcommand{\arraystretch}{1.15}
\begin{tabular}{lcccc}
\toprule
\textbf{Environment} & \textbf{\# Task Types} & \textbf{\# Train} & \textbf{\# Test} & \textbf{Source} \\
\midrule
ALFWorld     & 6   & 3{,}553 & 274 & \citet{shridhar2021alfworld} \\
ScienceWorld & 30  & 3{,}604 & 1{,}802 & \citet{wang2022scienceworld} \\
WebArena     & 241 & --      & 812 & \citet{zhou2024webarena} \\
\bottomrule
\end{tabular}
\caption{Train/test split statistics for the three interactive benchmark 
environments, using the original dataset releases. ALFWorld provides 
3{,}553 training games and a held-out \textit{seen} + \textit{unseen} test set of 274 
games across 6 compositional household task types~\citep{shridhar2021alfworld}. 
ScienceWorld contains 30 task types with 7{,}207 parametric variations 
in total, split 50\,\% / 25\,\% / 25\,\% into train / dev / test 
sets~\citep{wang2022scienceworld}; we report train and test only. 
WebArena is held-out only and not used during training; it consists of 
812 tasks instantiated from 241 intent templates across five self-hosted 
sites (Shopping, Shopping Admin/CMS, Reddit, GitLab, and 
Maps)~\citep{zhou2024webarena}.}
\label{tab:dataset_stats}
\end{table}

\paragraph{WebArena.}
WebArena consists of long-horizon web-navigation tasks served via a 
self-hosted, sandboxed deployment. We use 117 held-out tasks sampled 
from three task families---shopping (e-commerce product search and 
checkout), shopping\_admin (Magento admin operations), and gitlab 
(GitLab repository management). No WebArena trajectories are used for 
training; the consolidator trained on ScienceWorld is applied to 
WebArena without any additional updates, testing the cross-domain 
transfer claim of \S\ref{sec:exp_online}.

\FloatBarrier

\section{Impact Statement}
\label{appendix:impacts}
Auto-Dreamer is foundational research on long-term memory mechanisms for language agents. It does not introduce a new deployed system, a new dataset of human subjects, or a new generative capability tied to a specific application domain. All experiments are conducted in simulated agentic environments (ALFWorld, ScienceWorld, WebArena) that do not involve personal data or interaction with real users. As such, the immediate societal impact of this specific work is limited.

\FloatBarrier


\section{Online Memory-Construction Prompts}
\label{sec:appendix-online-prompts}

This appendix lists the system prompts used by the three roles in our online
memory pipeline: the per-trace \emph{writer}, the cross-trace
\emph{auto-dreamer} synthesizer, and the environment \emph{task agent}. We
report verbatim text from \texttt{opentinker/memory\_training/} and
\texttt{opentinker/environment/} after stripping a small number of defensive
phrases that we found contributed nothing to performance (these are clearly
marked below).

\paragraph{Shared task-agent prompt across baselines.}
The task-agent prompt (Sec.~\ref{sec:appendix-task-agent}) is held
\emph{identical} across every baseline (\texttt{no\_memory}, \texttt{writer\_only},
\texttt{reflexion}, \texttt{expel}, \texttt{awm}, \texttt{memp},
\texttt{reasoningbank}, \texttt{mem0}, \texttt{lightmem}, \texttt{auto\_dreamer},
\texttt{auto\_dreamer\_rl}). The only thing that differs across baselines is the
contents of the \texttt{=== Memory from past experience ===} block injected
at the end of the system prompt; the surrounding agent prompt is invariant.
Memory-aware variants of the agent prompt (e.g.~``treat memory as a hint, not
a recipe'') were tested in ablations and found to be net-negative on the
final aggregate score, so the reported runs use the un-instrumented agent
prompt.

\subsection{Writer Prompts}
\label{sec:appendix-writer}

The writer reads ONE episode trace (marked \textsc{Success} or \textsc{Fail})
and emits zero or more structured \texttt{INSERT\_SEMANTIC} or
\texttt{INSERT\_PROCEDURAL} blocks. Output is parsed verbatim into the bank.

\subsubsection{ALFWorld Writer}

\textbf{Success path.}
\begin{lstlisting}[breaklines=true,basicstyle=\ttfamily\footnotesize]
You are a Memory Agent. Read an episode trace from a task agent and
distill reusable knowledge into structured memory entries.

The task agent operates in ALFWorld -- a text-based household
environment where it must complete tasks like "put a clean apple
in the fridge" by issuing text commands (go to, take, clean, put,
etc.).

You will receive ONE episode trace, marked SUCCESS or FAIL.

Your output is injected into a task agent's system prompt to help
it succeed on NEW, unseen tasks of the same type.

OUTPUT FORMAT
You may emit one or more entries. Each entry must be one of:

INSERT_SEMANTIC
name: <short_id>
summary: <one-line description>
details: <full information>
END

INSERT_PROCEDURAL
name: <short_id>
type: <workflow|guide>
summary: <short description>
steps: ["step 1", "step 2", ...]
END

Nothing useful: NO_UPDATE

FORMAT RULES
- Use exact ALFWorld action names (go to, take, move, open, close,
  use, examine, heat, cool, clean, slice, inventory, look).
- Output entries then STOP.
\end{lstlisting}

\textbf{Failure path.}
\begin{lstlisting}[breaklines=true,basicstyle=\ttfamily\footnotesize]
[same header as Success]

A failed trace is useful for distilling the failure mode -- the
specific unproductive action pattern observed.

INSERT_SEMANTIC
name: <short_id>
summary: <failure mode actually observed>
details: concrete description of the failure mode in this trace,
quoting verbatim observations where possible.
END

If the trace is too short or noisy: NO_UPDATE
\end{lstlisting}

\subsubsection{ScienceWorld Writer}

\textbf{Success path.}
\begin{lstlisting}[breaklines=true,basicstyle=\ttfamily\footnotesize]
You are a Memory Agent. Read an episode trace and distill reusable
knowledge into structured memory entries.

The task agent operates in ScienceWorld -- a text-based scientific
reasoning environment with 30 distinct task types (measure-melting-
point, test-conductivity, grow-plant, chemistry-mix, find-animal,
mendelian-genetics, lifespan-*, inclined-plane-*, etc.). Each task
unfolds over multiple rooms (kitchen, workshop, greenhouse, art
studio, living room, bathroom, outside, foundry, bedroom, hallway).

The agent issues commands from templates such as:
    teleport to ROOM     open OBJ          pick up OBJ
    look at OBJ          look in OBJ       put down OBJ
    move OBJ to OBJ      pour OBJ in OBJ   dunk OBJ in OBJ
    mix OBJ              eat OBJ           use OBJ on OBJ
    activate OBJ         deactivate OBJ    flush OBJ
    connect OBJ to OBJ   disconnect OBJ    read OBJ
    focus on OBJ         wait              wait1

Your output is injected into a task agent's system prompt to help
it succeed on NEW, unseen variations of the SAME task type.

OUTPUT FORMAT
[INSERT_SEMANTIC / INSERT_PROCEDURAL blocks as in ALFWorld]

FORMAT RULES
- Use exact SciWorld action templates.
- Output entries then STOP.
\end{lstlisting}

\subsubsection{WebArena Writer}

\textbf{Success path.}
\begin{lstlisting}[breaklines=true,basicstyle=\ttfamily\footnotesize]
You are a Memory Agent. Read an episode trace from a web-browsing
task agent and distill reusable knowledge into structured memory
entries.

The task agent operates a real browser (Chromium, headless) via a
high-level action API. Each turn it receives the page's accessibility
tree (AXTree) as the observation and emits one action like
`click('123')`, `fill('42', 'value')`, `scroll(0, 200)`,
`keyboard_press('Enter')`, `goto(url)`, or `send_msg_to_user(text)`.

Your output is injected into a task agent's system prompt to help
it succeed on NEW tasks involving similar widgets or workflows.

OUTPUT FORMAT
[INSERT_SEMANTIC / INSERT_PROCEDURAL blocks as in ALFWorld; semantic
 entries describe widget knowledge, procedural describe action sequences]

FORMAT RULES
- Refer to elements by their VISIBLE LABEL or ROLE as it appears in
  the AXTree, NOT by element bid numbers (bids change every page load).
- Use exact action names: click, fill, scroll, keyboard_press, goto,
  select_option, hover, dblclick, drag_and_drop, send_msg_to_user.
- Output entries then STOP.
\end{lstlisting}

\textbf{Failure path.} On WebArena we observed that behavioral failure
entries (``submitted before verifying'', ``didn't wait for the page to
load'') generalize net-negative; the failure prompt explicitly excludes
them and only accepts concrete content/navigation mistakes:
\begin{lstlisting}[breaklines=true,basicstyle=\ttfamily\footnotesize]
A failed trace is useful ONLY when it shows a CONCRETE FACTUAL MISTAKE
the agent made -- a wrong field value, wrong navigation target, wrong
inferred number, wrong sub-page, wrong filter selection.

INSERT_SEMANTIC
name: <short_id>
summary: <one-line description of the specific content mistake>
details: which exact value / menu / page / number was wrong (e.g.
"agent picked Period 'Month' but goal asked for 'Year'", or
"agent navigated to Reports>Reviews when goal needed Reports>Bestsellers").
Quote the wrong action verbatim.
END

If the only failure-pattern is behavioural caution: NO_UPDATE
\end{lstlisting}

\subsection{Auto-Dreamer Synthesizer Prompt}
\label{sec:appendix-dreamer}
The same prompt is used across all environments.

\begin{lstlisting}[breaklines=true,basicstyle=\ttfamily\footnotesize]
You are a Memory Bank Synthesizer. Read a reference bank of memory
entries from past task sessions and create a compact, high-quality
output bank of synthesized entries. Only your synthesized entries
will be shown to the task agent.

The task agent uses these memories to succeed on NEW, unseen tasks.
Capture transferable knowledge -- patterns, procedures, and insights
that generalize across task instances. Entry summaries are used as
retrieval keys, so write clear, descriptive summaries.

Call ONE tool per turn. When satisfied, call `terminate`.

AVAILABLE TOOLS
Navigation (reference bank, read-only):
  search_memory(query, k=5)
  check_memory(ids=[...])    (up to 30 ids)
  get_source_trace(id)

Synthesis (output bank):
  synthesize(source_ids, type, name, summary, details?, steps?)

Control: terminate()

SYNTHESIS GUIDELINES
- Survey the reference bank broadly before synthesizing.
- Each entry should capture distinct, non-redundant knowledge.
- Look for patterns: shared procedures, recurring constraints,
  common strategies. Generalize when multiple entries support it.
- When entries disagree, resolve by frequency or note the conditions
  under which each applies; use get_source_trace to ground claims.
- Both procedural and semantic entries are valuable.
- Prefer actionable rules (priority "do X before Y", conditional
  "if Z, skip W") when well-supported.

GROUNDING RULES
- Every synthesized entry must cite source_ids drawn from the
  reference bank.
- Preserve concrete details that carry warning value: the exact
  forbidden command, the invalid action syntax the env rejected,
  the wrong object that ended an episode early.
\end{lstlisting}

\subsection{Task-Agent Prompts (shared across all baselines)}
\label{sec:appendix-task-agent}

The task agent is the policy that interacts with the environment.  All
baselines use the prompt below verbatim; the only difference between
\texttt{no\_memory} and the memory baselines is the presence of a
\texttt{=== Memory from past experience ===} block (followed by the
retrieved entries) appended to the system prompt at task start.

\subsubsection{ALFWorld Task Agent}
\begin{lstlisting}[breaklines=true,basicstyle=\ttfamily\footnotesize]
You are the Task Agent in an ALFWorld environment.
Your goal is to complete household tasks by executing actions.

CRITICAL RULE: Every observation includes "=== Available Actions ===".
You MUST pick EXACTLY ONE action from that list, word-for-word.
Any command not on the list will fail.

Memory (if provided below) describes high-level STRATEGIES in natural
language. It is NOT a list of executable commands. Use it to decide
WHICH action from the Available Actions list to pick, but always
output a command copied verbatim from the list.

Output ONLY a single action command.

Example response: go to desk 1
\end{lstlisting}

\subsubsection{ScienceWorld Task Agent}
\begin{lstlisting}[breaklines=true,basicstyle=\ttfamily\footnotesize]
You are a Task Agent playing ScienceWorld -- a text-based scientific
reasoning environment. Your goal is to complete the task described
to you by issuing text commands, one per turn.

NAVIGATION -- USE TELEPORT. Rooms are connected by doors that may be
closed. Instead of "go door to kitchen" (which often fails), always
prefer:
    teleport to kitchen / teleport to workshop / teleport to greenhouse
    teleport to hallway  / teleport to bedroom  / teleport to art studio
    teleport to living room / teleport to bathroom / teleport to outside
    teleport to foundry
Teleport always succeeds and is always available.

ACTION TEMPLATES (any action emitted must match one):
  teleport to ROOM    go OBJ              look around
  look at OBJ         look in OBJ         inventory
  open OBJ            close OBJ           pick up OBJ
  put down OBJ        move OBJ to OBJ     pour OBJ in OBJ
  dunk OBJ in OBJ     mix OBJ             eat OBJ
  read OBJ            use OBJ on OBJ
  activate OBJ        deactivate OBJ      flush OBJ
  connect OBJ to OBJ  disconnect OBJ
  focus on OBJ        wait                wait1

OBJECTS: each turn shows a "Visible objects" list -- those are the
base names the env recognises right now. Compound names like
"substance in metal pot" are also accepted. When an object is hidden
inside a closed container, first `open` or `look in` to reveal it.

DISAMBIGUATION: when an action targets an object with multiple
instances, the env replies "Ambiguous request: please enter the
number..." -- emit the corresponding number on the next turn.
\end{lstlisting}

\subsubsection{WebArena Task Agent}
\begin{lstlisting}[breaklines=true,basicstyle=\ttfamily\footnotesize]
You are a web-browsing agent. Each turn, observe the AXTree and emit
EXACTLY ONE action in a fenced block:

```action
click('123')
```

Use `bid` strings (e.g. [42]) from the AXTree to refer to elements.
Emit `send_msg_to_user('answer')` when done.

Action space:
{action_desc}    % auto-injected: 12 BrowserGym high-level actions
                 % (click, fill, scroll, keyboard_press, goto,
                 %  select_option, hover, dblclick, drag_and_drop,
                 %  noop, send_msg_to_user)
\end{lstlisting}

\FloatBarrier

\section{Usage of LLMs}
\label{appendix:llm_usage}

We used LLMs as a writing assistant to help us edit parts of the paper. Additionally, we utilize the power of CodeX and Claude Code to help us code faster. All AI-generated writing and code are manually checked and modified. There is no fully AI-generated content in the paper.

\FloatBarrier

\section{Additional Per-Baseline Memory-Construction Prompts}
\label{app:baseline-prompts}

Reproduced verbatim from \texttt{baselines/*/memory.py}, which re-read
each baseline's reference repo at the commit hash recorded in the
file's top docstring. Long verbatim few-shot examples are replaced by
a \texttt{[\ldots]} marker citing the precise file:line.

\paragraph{ReasoningBank.}
Successful-trajectory extraction (Tab.~\ref{tab:prompt-rb-success}),
failed-trajectory extraction (Tab.~\ref{tab:prompt-rb-failure}), and
the memory-injection banner prepended to retrieved items at inference
time (Tab.~\ref{tab:prompt-rb-banner}).

\begin{table}[!htbp]
\centering\small
\caption{\textbf{ReasoningBank --- successful-trajectory extraction prompt}
}
\label{tab:prompt-rb-success}
\begin{tabular}{p{13cm}}
\toprule[1.1pt]
You are an expert in household environment navigation. You will be given a user query, the corresponding trajectory that represents \textbf{how an agent successfully accomplished the task}.\\[4pt]
\textbf{\#\# Guidelines}\\
You need to extract and summarize useful insights in the format of memory items based on the agent's successful trajectory.\\
The goal of summarized memory items is to be helpful and generalizable for future similar tasks.\\[4pt]
\textbf{\#\# Important notes}\\
- You must first think why the trajectory is successful, and then summarize the insights.\\
- You can extract \emph{at most 3} memory items from the trajectory.\\
- You must not repeat similar or overlapping items.\\
- Prefer concrete, actionable procedures over abstract principles. Do not embed specific product names, queries, or literal string contents from the task.\\[4pt]
\textbf{\#\# Output Format}\\
Your output must strictly follow the Markdown format shown below:\\[4pt]
\texttt{\# Memory Item i}\\
\texttt{\#\# Title <the title of the memory item>}\\
\texttt{\#\# Description <one sentence summary describing when or when NOT to use the memory item>}\\
\texttt{\#\# Content <1-3 sentences describing the insights learned to successfully accomplishing similar tasks in the future>}\\
\bottomrule[1.1pt]
\end{tabular}
\end{table}

\begin{table}[!htbp]
\centering\small
\caption{\textbf{ReasoningBank - failed-trajectory extraction prompt}}
\label{tab:prompt-rb-failure}
\begin{tabular}{p{13cm}}
\toprule[1.1pt]
You are an expert in household environment navigation. You will be given a user query, the corresponding trajectory that represents \textbf{how an agent attempted to resolve the task but failed}.\\[4pt]
\textbf{\#\# Guidelines}\\
You need to extract and summarize useful insights in the format of memory items based on the agent's failed trajectory.\\
The goal of summarized memory items is to be helpful and generalizable for future similar tasks.\\[4pt]
\textbf{\#\# Important notes}\\
- You must first reflect and think why the trajectory failed, and then summarize what lessons you have learned or strategies to prevent the failure in the future.\\
- You can extract \emph{at most 3} memory items from the trajectory.\\
- You must not repeat similar or overlapping items.\\
- Prefer concrete, actionable recovery procedures over abstract principles. Do not embed specific product names, queries, or literal string contents from the task.\\[4pt]
\textbf{\#\# Output Format} (same Markdown schema as Table~\ref{tab:prompt-rb-success}, with ``successfully accomplishing'' replaced by ``avoid such failures and successfully accomplishing'').\\
\bottomrule[1.1pt]
\end{tabular}
\end{table}

\begin{table}[!htbp]
\centering\small
\caption{\textbf{ReasoningBank --- memory-injection banner prepended to retrieved memory items at inference time}}
\label{tab:prompt-rb-banner}
\begin{tabular}{p{13cm}}
\toprule[1.1pt]
Below are some memory items that I accumulated from past interaction from the environment that may be helpful to solve the task. You can use it when you feel it's relevant. In each step, please first explicitly discuss if you want to use each memory item or not, and then take action.\\
\bottomrule[1.1pt]
\end{tabular}
\end{table}

\paragraph{ExpeL.}
A rule-operation grammar (Tab.~\ref{tab:prompt-expel-rules}) appended
to the compare-critique template (Tab.~\ref{tab:prompt-expel-compare},
fired on each success/failure pair) and the all-success template
(Tab.~\ref{tab:prompt-expel-allsucc}, fired every $8$ successes).

\begin{table}[!htbp]
\centering\small
\caption{\textbf{ExpeL --- rule-operation format template}}
\label{tab:prompt-expel-rules}
\begin{tabular}{p{13cm}}
\toprule[1.1pt]
\texttt{<OPERATION> <RULE NUMBER>: <RULE>}\\[4pt]
The available operations are: AGREE (if the existing rule is strongly relevant for the task), REMOVE (if one existing rule is contradictory or similar/duplicated to other existing rules), EDIT (if any existing rule is not general enough or can be enhanced, rewrite and improve it), ADD (add new rules that are very different from existing rules and relevant for other tasks). Each needs to CLOSELY follow their corresponding formatting below (any existing rule not edited, not agreed, nor removed is considered copied):\\[4pt]
\texttt{AGREE <EXISTING RULE NUMBER>: <EXISTING RULE>}\\
\texttt{REMOVE <EXISTING RULE NUMBER>: <EXISTING RULE>}\\
\texttt{EDIT <EXISTING RULE NUMBER>: <NEW MODIFIED RULE>}\\
\texttt{ADD <NEW RULE NUMBER>: <NEW RULE>}\\[4pt]
Do not mention the trials in the rules because all the rules should be GENERALLY APPLICABLE. Each rule should be concise and easy to follow. Any operation can be used MULTIPLE times. Do at most 4 operations and each existing rule can only get a maximum of 1 operation.\\
\bottomrule[1.1pt]
\end{tabular}
\end{table}

\begin{table}[!htbp]
\centering\small
\caption{\textbf{ExpeL --- compare-critique prompt}}
\label{tab:prompt-expel-compare}
\begin{tabular}{p{13cm}}
\toprule[1.1pt]
\texttt{\{instruction\}}\\
Here are the two previous trials to compare and critique:\\
\textbf{TRIAL TASK}: \texttt{\{task\}}\\
\textbf{SUCCESSFUL TRIAL}: \texttt{\{success\_history\}}\\
\textbf{FAILED TRIAL}: \texttt{\{fail\_history\}}\\
\textbf{Here are the EXISTING RULES}: \texttt{\{existing\_rules\}}\\[4pt]
By examining and contrasting to the successful trial, and the list of existing rules, you can perform the following operations: add, edit, remove, or agree so that the new list of rules is GENERAL and HIGH LEVEL critiques of the failed trial or proposed way of Thought so they can be used to avoid similar failures when encountered with different questions in the future. Have an emphasis on critiquing how to perform better Thought and Action. Follow the below format:\\[4pt]
[Rule-operation format from Table~\ref{tab:prompt-expel-rules} appended verbatim.]\\
\bottomrule[1.1pt]
\end{tabular}
\end{table}

\begin{table}[!htbp]
\centering\small
\caption{\textbf{ExpeL --- all-success critique prompt}}
\label{tab:prompt-expel-allsucc}
\begin{tabular}{p{13cm}}
\toprule[1.1pt]
\texttt{\{instruction\}}\\
Here are the trials: \texttt{\{success\_history\}}\\
Here are the EXISTING RULES: \texttt{\{existing\_rules\}}\\[4pt]
By examining the successful trials, and the list of existing rules, you can perform the following operations: add, edit, remove, or agree so that the new list of rules are general and high level insights of the successful trials or proposed way of Thought so they can be used as helpful tips to different tasks in the future. Have an emphasis on tips that help the agent perform better Thought and Action. Follow the below format:\\[4pt]
[Rule-operation format from Table~\ref{tab:prompt-expel-rules} appended verbatim.]\\
\bottomrule[1.1pt]
\end{tabular}
\end{table}

\paragraph{LightMem.}
STM$\to$LTM extraction (Tab.~\ref{tab:prompt-lightmem-extract}) and
offline UPDATE/DELETE/IGNORE consolidation
(Tab.~\ref{tab:prompt-lightmem-update}).

\begin{table}[!htbp]
\centering\small
\caption{\textbf{LightMem --- STM$\to$LTM extraction prompt}}
\label{tab:prompt-lightmem-extract}
\begin{tabular}{p{13cm}}
\toprule[1.1pt]
You are a Personal Information Extractor.\\
Your task is to extract \textbf{all possible facts or information} about the user from a conversation, where the dialogue is organized into topic segments separated by markers like:\\[4pt]
Input format: \texttt{--- Topic X ---}; \texttt{[timestamp, weekday] source\_id.SpeakerName: message}\\[4pt]
\textbf{Important Instructions}:\\
0. You MUST process messages \emph{strictly in ascending sequence\_number order}. For each message, stop and carefully evaluate before moving to the next. Do NOT reorder, batch-skip, or skip ahead.\\
1. You MUST process every user message in order. For each, decide whether it contains factual information; if yes extract and rephrase as a standalone sentence; if no (pure greeting/filler) skip. Do NOT skip just because it looks minor.\\
2. Perform light contextual completion so each fact is a standalone statement.\\
3. Use the \texttt{sequence\_number} (integer prefix before each message) as the \texttt{source\_id}.\\
4. Output as JSON: \texttt{\{"data": [\{"source\_id": <id>, "fact": "<complete fact>"\}]\}}.\\[4pt]
\textbf{Reminder}: Be exhaustive. Unless a message is purely meaningless, extract and output it as a fact.\\
\bottomrule
\end{tabular}
\end{table}

\begin{table}[!htbp]
\centering\small
\caption{\textbf{LightMem --- offline UPDATE/DELETE/IGNORE consolidation prompt}}
\label{tab:prompt-lightmem-update}
\begin{tabular}{p{13cm}}
\toprule[1.1pt]
You are a memory management assistant. Your task is to decide whether the target memory should be updated, deleted, or ignored based on the candidate source memories.\\[4pt]
\textbf{Decision rules}:\\
1. \emph{Update}: target and candidates describe essentially the same fact/event but are not fully consistent (candidates provide more details, refinements, or clarifications) $\to$ update by integrating the additional information.\\
2. \emph{Delete}: target and candidates contain a direct conflict; candidates (more recent) take precedence $\to$ delete the target.\\
3. \emph{Ignore}: target and candidates are unrelated $\to$ no action.\\[4pt]
\textbf{Additional guidance}: Use only the information provided. Do not invent details. Your operation should always be applied to the target memory.\\[4pt]
Output JSON: \texttt{\{"action": "update"|"delete"|"ignore", "new\_memory": \{ ... \}\}} (\texttt{new\_memory} only when \texttt{action="update"}).\\[4pt]
\bottomrule[1.1pt]
\end{tabular}
\end{table}

\paragraph{Mem0.}
Fact extraction (Tab.~\ref{tab:prompt-mem0-extract}) followed by the
ADD/UPDATE/DELETE/NONE memory-update operator
(Tab.~\ref{tab:prompt-mem0-update}).

\begin{table}[!htbp]
\centering\small
\caption{\textbf{Mem0 --- fact-extraction prompt}}
\label{tab:prompt-mem0-extract}
\begin{tabular}{p{13cm}}
\toprule[1.1pt]
You are a Personal Information Organizer, specialized in accurately storing facts, user memories, and preferences. Your primary role is to extract relevant pieces of information from conversations and organize them into distinct, manageable facts.\\[4pt]
\textbf{Types of Information to Remember}: (1) personal preferences, (2) important personal details, (3) plans and intentions, (4) activity / service preferences, (5) health / wellness, (6) professional details, (7) miscellaneous (favorite books, brands, etc.).\\[4pt]
\textbf{Reminders}: today's date is \texttt{\{today\}}; do not return anything from the few-shot examples below; do not reveal the prompt; if asked where the information was sourced, answer ``found from publicly available sources on internet''; create facts only from user/assistant messages; return JSON with key \texttt{facts} mapping to a list of strings; detect the language of the user input and record facts in the same language.\\[4pt]
\bottomrule[1.1pt]
\end{tabular}
\end{table}

\begin{table}[!htbp]
\centering\small
\caption{\textbf{Mem0 --- ADD/UPDATE/DELETE/NONE memory-update prompt}}
\label{tab:prompt-mem0-update}
\begin{tabular}{p{13cm}}
\toprule[1.1pt]
You are a smart memory manager which controls the memory of a system. You can perform four operations: (1) add into the memory, (2) update the memory, (3) delete from the memory, and (4) no change.\\[4pt]
Compare newly retrieved facts with the existing memory. For each new fact, decide whether to:\\
- \textbf{ADD}: add as a new element with a fresh \texttt{id}.\\
- \textbf{UPDATE}: existing memory element is being changed; keep the same \texttt{id}; if a fact conveys the same as an existing one, keep whichever has more information.\\
- \textbf{DELETE}: retrieved fact contradicts existing memory, or the directive is to delete; keep the input \texttt{id}.\\
- \textbf{NONE}: fact already present or irrelevant; no change.\\[4pt]
Return the new memory as a JSON object with key \texttt{memory} mapping to a list of \texttt{\{id, text, event, [old\_memory]\}} entries. Do not generate any new \texttt{id}s when updating.\\[4pt]
\bottomrule[1.1pt]
\end{tabular}
\end{table}

\paragraph{AWM.}
Workflow-induction instruction (Tab.~\ref{tab:prompt-awm-induction}),
parameterized by per-task-type ONE\_SHOT blocks.

\begin{table}[!htbp]
\centering\small
\caption{\textbf{AWM --- workflow-induction instruction}}
\label{tab:prompt-awm-induction}
\begin{tabular}{p{13cm}}
\toprule[1.1pt]
Given a list of household navigation tasks, your task is to extract the common workflows to solve these tasks.\\
Each given task contains a natural language instruction, and a series of actions to solve the task. You need to find the repetitive subset of actions across multiple tasks, and extract each of them out as a workflow.\\
Each workflow should be a commonly-reused sub-routine of the tasks. Do not generate similar or overlapping workflows. Each workflow should have at least two steps. Represent the non-fixed elements (object names, receptacle ids) with descriptive variable names as shown in the example.\\
Keep the values of invariant elements, e.g., the literal verb ``heat'' or ``cool'', as they will share and stay invariant across tasks.\\
Try to generate as many workflows that can cover all the tasks in the input list.\\[4pt]
{[\ldots followed by a per-task-type ONE\_SHOT block (one Concrete Examples worked trajectory $+$ one Summary Workflows block per ALFWorld task family); full per-type ONE\_SHOTs at \texttt{baselines/awm/memory.py:167--360}\ldots]}\\
\bottomrule[1.1pt]
\end{tabular}
\end{table}

\paragraph{Memp.}
Workflow construction (Tab.~\ref{tab:prompt-memp-build}) and
failure-driven workflow adjustment
(Tab.~\ref{tab:prompt-memp-adjust}).

\begin{table}[!htbp]
\centering\small
\caption{\textbf{Memp --- workflow-build prompt}
}
\label{tab:prompt-memp-build}
\begin{tabular}{p{13cm}}
\toprule[1.1pt]
You are provided with a query and a trajectory taken to solve the query. The trajectory consists of multiple steps of thought, action and observation.\\
Your task is to generate a workflow based on critical steps to help solve similar queries in the future.\\
A critical step is one that has a significant impact on fulfilling the query, the step action belongs to the set [\texttt{go to, take from, put in/on, open, close, use, clean with, heat with, cool with, examine, look}], and the action's outcome is successful and contributes positively to achieving the query.\\
\textbf{Notice}: Write the workflow as a natural, coherent paragraph (not as a bullet list or numbered steps). Use clear, concise language to describe what actions should be taken and in what general order.\\[4pt]
\textbf{-----EXAMPLE WORKFLOW----}\\
To solve this query, begin by identifying the most likely receptacles where the target object can be found and visit them one by one. After locating and taking the object, perform any required transformation such as cleaning at a sinkbasin, heating with a microwave, or cooling with a fridge. Finally, go to the destination receptacle and put the object in/on it to complete the task.\\
\textbf{-----EXAMPLE END----}\\[4pt]
\textbf{Query}: \texttt{\{query\}}\\
\textbf{Trajectory}: \texttt{\{trajectory\}}\\
- \textbf{DO NOT} copy \texttt{Thought:}, \texttt{Action:}, or \texttt{Observation:} lines from the trajectory above.\\
Output the workflow without any explanation or context:\\
\bottomrule[1.1pt]
\end{tabular}
\end{table}

\begin{table}[!htbp]
\centering\small
\caption{\textbf{Memp --- failure-Adjustment prompt}
}
\label{tab:prompt-memp-adjust}
\begin{tabular}{p{13cm}}
\toprule[1.1pt]
You are a helpful assistant. You are given a workflow, a reward, and a trajectory.\\
Reward is a number between 0 and 1; 1 means the trajectory is successful, 0 means failed.\\
If the reward is False (i.e., the trajectory guided by the workflow did not successfully complete the task), then analyze why the task was not completed based on the trajectory and the workflow.\\
After that, refine the workflow to make it more accurate and robust, so that it can better guide the completion of the task.\\[4pt]
\textbf{Workflow}: \texttt{\{workflow\}}\\
\textbf{Reward}: \texttt{\{reward\}}\\
\textbf{Trajectory}: \texttt{\{trajectory\}}\\[4pt]
{[\ldots six verbatim per-task-type Output Example Workflows (one each for \texttt{pick\_and\_place}, \texttt{pick\_clean\_then\_place}, \texttt{pick\_heat\_then\_place}, \texttt{pick\_cool\_then\_place}, \texttt{look\_at\_obj}, \texttt{pick\_two\_obj}) omitted; full prompt at \texttt{baselines/memp/memory.py:121--127}\ldots]}\\[4pt]
Keep your output in the format below:\\
\texttt{<Analysis> your analysis here </Analysis>}\\
\texttt{<Workflow> your adjusted workflow here </Workflow>}\\
\bottomrule[1.1pt]
\end{tabular}
\end{table}

\paragraph{Reflexion.}
Reflector prompt fired on failure only
(Tab.~\ref{tab:prompt-reflexion-reflector}).

\begin{table}[!htbp]
\centering\small
\caption{\textbf{Reflexion --- failure-reflection prompt}}
\label{tab:prompt-reflexion-reflector}
\begin{tabular}{p{13cm}}
\toprule[1.1pt]
You are a self-reflection agent. The agent attempted the following task and FAILED.\\
\textbf{Task}: \texttt{\{task\}}\\
\textbf{Trajectory}: \texttt{\{traj\}}\\[4pt]
Write a short paragraph (at most 4 sentences) reflecting on what went wrong and a concrete strategy to try next time on a SIMILAR task. Be specific and actionable.\\[4pt]
Reflection:\\
\bottomrule[1.1pt]
\end{tabular}
\end{table}

\FloatBarrier

\section{Bootstrap Confidence Intervals}
\label{appx:significance}

We complement the point estimates in Table~\ref{tab:main_memory_eval} with
bootstrap 95\% confidence intervals on per-method success rate. For each
domain we resample tasks with replacement ($N_B = 10{,}000$) and report the
bootstrap mean and 95\% percentile interval. WebArena CIs are computed on
the per-task-family macro average to match the main-text metric.

\begin{table}[!htbp]
\centering
\small
\setlength{\tabcolsep}{4pt}
\renewcommand{\arraystretch}{1.05}
\caption{Bootstrap 95\% CIs on continual-memory deployment success rate
($N_B = 10{,}000$). ScienceWorld and ALFWorld are evaluated with per-task
SR; WebArena uses per-task-family macro-averaged SR over 117 tasks across
shopping, shopping\_admin, and gitlab.}
\label{tab:bootstrap_cis}
\begin{tabular}{l ccc}
\toprule
\textbf{Method} 
& \textbf{ScienceWorld} 
& \textbf{ALFWorld} 
& \textbf{WebArena} \\
\midrule
No memory      & 28.69 [25.18, 32.14] & 30.83 [26.40, 35.31] & 44.6 [36.0, 53.2] \\
Reflexion      & 29.55 [26.07, 33.10] & 49.17 [43.59, 54.73] & 46.4 [37.7, 54.9] \\
ExpeL          & 28.33 [24.97, 31.76] & 33.59 [28.79, 38.48] & 50.8 [41.8, 59.7] \\
AWM            & 30.18 [26.61, 33.72] & 32.75 [28.16, 37.47] & 52.0 [43.1, 60.8] \\
Memp           & 30.36 [26.90, 33.96] & 31.40 [27.09, 35.76] & 50.8 [41.7, 59.7] \\
ReasoningBank  & 30.86 [27.32, 34.55] & 31.11 [26.56, 35.84] & 49.8 [41.1, 58.6] \\
Mem0           & 26.79 [23.39, 30.21] & 30.61 [25.81, 35.42] & 50.3 [41.5, 59.2] \\
LightMem       & 28.08 [23.90, 32.37] & 31.16 [26.72, 35.53] & 52.0 [43.1, 60.6] \\
Mem-$\alpha$   & 29.97 [26.43, 33.51] & 57.40 [52.04, 62.59] & --- \\
UMEM           & 34.07 [30.51, 37.71] & 58.43 [53.15, 63.65] & --- \\
\midrule
\rowcolor{hl-ours-a}
\textbf{Auto-Dreamer} 
& \textbf{41.07 [37.53, 44.70]}
& \textbf{60.21 [54.65, 65.59]}
& \textbf{52.3 [43.5, 60.9]} \\
\bottomrule
\end{tabular}
\end{table}

\FloatBarrier

\section{Case Study: LightMem vs Auto-Dreamer}
\label{sec:case-study-lifespan}

We include a qualitative case study to illustrate how Auto-Dreamer achieves compactness without sacrificing task success. We run Auto-Dreamer and LightMem on the same 96 episodes from the ScienceWorld \texttt{lifespan-compare} category, using the same task order, task agent, writer, retriever, and memory-token budget. Both methods solve 48 out of 96 tasks, yielding identical success rate of $50.0\%$. The difference is therefore not task accuracy in this slice, but the structure and size of the memory bank that supports future retrieval.

At episode~90, LightMem has 265 active entries totaling 17{,}512 tokens, whereas Auto-Dreamer has 14 active entries totaling 716 tokens. Thus, in this run, Auto-Dreamer maintains a $24.5\times$ smaller active bank in tokens and an $18.9\times$ smaller bank in entry count while matching LightMem's task success.

\begin{center}
\small
\captionof{table}{Case study on the ScienceWorld \texttt{lifespan-compare} category. Auto-Dreamer matches LightMem's success while maintaining a substantially smaller active memory bank.}
\label{tab:case-study-lifespan}
\begin{tabular}{lccc}
\toprule
Method & Success & Active entries at $t{=}90$ & Active tokens at $t{=}90$ \\
\midrule
LightMem & $48/96$ ($50.0\%$) & 265 & 17{,}512 \\
Auto-Dreamer & $48/96$ ($50.0\%$) & 14 & 716 \\
\bottomrule
\end{tabular}
\end{center}

\paragraph{LightMem accumulates surface-level duplicates.}
LightMem's bank is dominated by near-verbatim restatements of task instructions and local state observations. Among the 265 active entries at $t{=}90$, 49 are paraphrases of the task instruction. The first four such entries are byte-identical:
\begin{quote}\small\itshape
``The task is to find the animal with the longest life span, then the shortest life span. The animals are located in the `outside' location. Additionally, there are sequential subgoals to focus on the animal with the shortest life span.''
\end{quote}
The bank also contains repeated trivial state shards, including four byte-identical copies of \textit{``The agent's inventory contains an orange.''} and two copies of \textit{``The agent has taken 0 moves so far.''} It further stores multiple reorderings of the same room description:
\begin{quote}\small\itshape
[ltm\#32]: ``In the foundry, there is a blast furnace that is turned off and has a closed door, a sink that is turned off and contains nothing, a table that contains nothing, and a door to the outside that is open.''\\
{}[ltm\#33]: ``In the foundry, there is a sink that is turned off and contains nothing, a blast furnace that is turned off and has a closed door, a table that contains nothing, and a door to the outside that is open.''\\
{}[ltm\#34, \#35]: \textit{further reorderings of the same four objects.}
\end{quote}
The remaining bank includes many long summaries that recapitulate the full visited world state together with histories of invalid actions. In this run, LightMem's consolidation step fires nine times but retires no active entries, so memory grows monotonically.

\paragraph{Auto-Dreamer replaces instances with abstractions.}
Auto-Dreamer's first consolidation trigger fires at episode~9. At that point, the writer has emitted candidate memories from five trajectories, each tied to a different concrete focus target: \textit{crocodile}, \textit{egg giant tortoise}, \textit{baby brown bear}, \textit{baby elephant}, and a generic \textit{animal}. Rather than preserving these as separate task-specific entries, Auto-Dreamer collapses them into a single procedural rule:
\begin{quote}\small\itshape
\textbf{General procedure for lifespan comparison tasks:}\\
- teleport to outside\\
- focus on animal (\textbf{prefer adult over juvenile or egg})
\end{quote}
This entry preserves the reusable structure shared across the five trajectories while omitting episode-specific target names that would compete at retrieval time.

Auto-Dreamer also synthesizes memories that abstract recurring failure modes. For example, from failed trajectories in which the agent focused on \textit{baby baby beaver} and \textit{chameleon egg}, it writes:
\begin{quote}\small\itshape
\textbf{Common incorrect targets in lifespan comparison tasks:}
Focusing on juveniles or eggs instead of adult animals can lead to failure in lifespan comparison tasks.
\end{quote}
A later consolidation event, using five additional failed trajectories from episodes 10--17, generalizes the same pattern into a higher-level procedural memory about focus-target accuracy. Together, the retained entries cover three complementary facts: where relevant animals are located, which targets are usually incorrect, and how to choose the correct adult target.

\paragraph{Retrieval becomes less redundant.}
The effect is visible at evaluation time. On \texttt{lifespan-shortest-lived::119} at episode~90, both methods succeed. However, the top retrieved memories differ sharply. LightMem retrieves the same task-instruction sentence three times, each from a different timestamp. Auto-Dreamer retrieves three distinct pieces of information: the relevant location, a common anti-pattern, and the success criterion for selecting the target. Thus, even when both agents solve the task, Auto-Dreamer uses each retrieval slot to expose a different abstraction, whereas LightMem spends multiple slots on duplicated content.

\paragraph{Why this case matters.}
This example illustrates the mechanism behind Auto-Dreamer's memory-efficiency gains. LightMem's prompted consolidation is conservative: it tolerates near-duplicates that differ only in surface form and does not reliably merge repeated observations into higher-level rules. Auto-Dreamer instead treats consolidation as region rewriting: it retires multiple concrete memories and replaces them with fewer provenance-grounded abstractions. In this case, LightMem grows approximately linearly with episodes, reaching 265 active entries by episode~90, whereas Auto-Dreamer reaches a compact steady state of roughly 10--14 active entries within the first half of the stream.

Overall, on \texttt{lifespan-compare}, both systems achieve the same $50.0\%$ success rate over 96 episodes, but Auto-Dreamer maintains a $24.5\times$ smaller active memory bank in tokens. This supports the main quantitative finding that learned offline consolidation improves the success--cost tradeoff not by merely deleting memories, but by replacing redundant local observations with compact, reusable abstractions.

\FloatBarrier

\section{Case Study: AWM vs Auto-Dreamer}
\label{sec:case-study-find-entity}

We include a second case study to separate memory compactness from memory usefulness. We compare AWM and Auto-Dreamer on the ScienceWorld \texttt{find-entity} category under the same online evaluation stream. Both methods use the same task agent, writer, retriever, prompt format, and task order; they differ only in the long-horizon memory that is retained and consolidated. AWM induces a compact library of natural-language workflow templates from successful trajectories, whereas Auto-Dreamer rewrites writer-emitted memories into synthesized semantic and procedural lessons.

After 80 capped episodes, the two methods have nearly identical active-bank footprints: AWM stores 10 entries totaling 816 tokens, while Auto-Dreamer stores 13 entries totaling 795 tokens. The performance gap is nevertheless large: Auto-Dreamer solves 58 out of 80 tasks ($72.5\%$), compared with 21 out of 80 for AWM ($26.2\%$). Successful Auto-Dreamer episodes are also shorter, averaging 7.4 steps compared with 20.0 steps for AWM. Thus, in this category, the advantage is not explained by a larger memory bank, but by what the bank contains.

\begin{center}
\small
\captionof{table}{Case study on ScienceWorld \texttt{find-entity}. AWM and Auto-Dreamer end the 80-episode stream with nearly identical active memory footprints, but Auto-Dreamer achieves substantially higher success and shorter successful trajectories.}
\label{tab:case-study-find-entity}
\begin{tabular}{lcccc}
\toprule
Method & Success & Avg. steps on successes & Active entries at $t{=}80$ & Active tokens at $t{=}80$ \\
\midrule
AWM & $21/80$ ($26.2\%$) & 20.0 & 10 & 816 \\
Auto-Dreamer & $58/80$ ($72.5\%$) & 7.4 & 13 & 795 \\
\bottomrule
\end{tabular}
\end{center}

\paragraph{AWM stores compact but underspecified workflows.}
At episode~80, AWM's memory consists of a single \texttt{find-entity} workflow bucket with 10 entries. These entries are concise action templates, such as:
\begin{quote}\small\itshape
\textbf{Workflow: Locate \texttt{\{object\}} by scanning likely receptacles}\\
\textbf{think:} An \texttt{\{object\}} is more likely to appear in \texttt{\{receptacle\_list\}}. Check candidates one by one until found.\\
\textbf{act:} go to \texttt{\{receptacle\_1\}}\\
\textbf{act:} open \texttt{\{receptacle\_1\}}\\
\textbf{act:} go to \texttt{\{receptacle\_2\}}\\
\textbf{act:} open \texttt{\{receptacle\_2\}}
\end{quote}
and:
\begin{quote}\small\itshape
\textbf{Workflow: Focus and pick up \texttt{\{object\}}}\\
\textbf{think:} The \texttt{\{object\}} is visible and can be picked up.\\
\textbf{act:} focus on \texttt{\{object\}}\\
\textbf{act:} pick up \texttt{\{object\}}
\end{quote}
The bank also contains several near-paraphrases of a canonical recipe:
\begin{quote}\small\itshape
focus on \texttt{\{object\}} $\rightarrow$ pick up \texttt{\{object\}} $\rightarrow$ teleport to \texttt{\{target\_location\}} $\rightarrow$ move \texttt{\{object\}} to \texttt{\{target\_receptacle\}}.
\end{quote}

These workflows are compact, but they primarily encode how to act once the target has already been identified. They do not encode the key semantic precondition for \texttt{find-entity}: the visible objects in the spawn room are often distractors, and the target entity is frequently elsewhere. In particular, the AWM bank contains no negative rule saying not to focus on salient but incorrect objects such as a banana, orange, apple, refrigerator, or bee hive before identifying the requested entity.

\paragraph{Auto-Dreamer stores identification knowledge and negative lessons.}
Auto-Dreamer's bank at episode~80 has 13 entries: 10 synthesized by the dreamer and 3 fresh writer entries that have not yet been consolidated. Its highest-leverage entries include both procedural and semantic abstractions. A representative synthesized procedural entry is:
\begin{quote}\small\itshape
\textbf{\texttt{find-entity-general-procedure-synthesized}}\\
- teleport to outside area\\
- focus on the entity\\
- pick up the entity\\
- teleport to the destination room\\
- move the entity to the specified container
\end{quote}
This entry is synthesized from five retired writer notes, with provenance tracing back to 22 originating trajectories. Its predecessors include concrete variants such as moving an entity to a yellow box, a blue box, or another destination container. Auto-Dreamer therefore collapses many task-specific action traces into one reusable procedure.

More importantly, Auto-Dreamer also stores a negative semantic lesson that AWM's success-only workflow induction does not produce:
\begin{quote}\small\itshape
\textbf{\texttt{avoid-wrong-focus}}\\
The agent should avoid focusing on objects that are not the target entity before completing the task setup. Focusing on incorrect objects, such as a bee hive, orange, or apple, leads to early failure. It is crucial to identify the correct target entity first before performing any interaction actions.
\end{quote}
This short entry is synthesized from failed trajectories and remains active through episode~80. The resulting bank covers four complementary facets of the task: where to search first, what objects to avoid, the canonical pick-and-move procedure, and common destination containers.

\paragraph{A side-by-side episode.}
Consider task \texttt{find-animal::254} at episode~29:
\begin{quote}\small\itshape
``Your task is to find a(n) animal. First, focus on the thing. Then, move it to the blue box in the living room.''
\end{quote}
Both agents start in the same kitchen. The opening observation includes a counter containing a bowl with a red apple, a banana, an orange, and a potato, along with closed storage objects such as a cupboard, freezer, and fridge.

The AWM agent receives its retrieved workflow templates and immediately executes:
\begin{quote}\small\itshape
\texttt{focus on banana}
\end{quote}
The episode terminates after one step with final score $-1.0$. The failure is consistent with the contents of the AWM bank: several workflows begin with \texttt{focus on \{object\}}, but none encode that a banana is not an animal or that the agent should first leave the misleading spawn location.

The Auto-Dreamer agent retrieves three compact entries, including facts that the entity is typically outside and should ultimately be moved to the living room. It then follows the successful trajectory:
\begin{quote}\small\itshape
\texttt{teleport to outside}\\
\texttt{focus on common toad}\\
\texttt{pick up common toad}\\
\texttt{teleport to living room}\\
\texttt{move common toad to blue box}
\end{quote}
The episode completes successfully in five steps. The task, starting observation, task agent, and writer are the same; the difference is the long-horizon memory exposed to the agent.

\paragraph{Why AWM struggles on \texttt{find-entity}.}
AWM is well matched to settings where successful demonstrations share a stable action skeleton. However, \texttt{find-entity} requires both procedural knowledge and identification knowledge. The agent must decide what counts as the requested entity, avoid salient distractors, and often navigate away from the initial room before interacting. A compact action template cannot express these preconditions unless they appear explicitly in the induced workflow.

The logs support this interpretation. Across the 80 capped tasks, AWM's first action is a \texttt{focus on ...} command in 20 episodes, and 11 of those episodes terminate immediately with score $-1.0$ after focusing on an incorrect object. These failures do not enter AWM's success-derived workflow pool. Auto-Dreamer, in contrast, can consolidate writer notes from failed episodes and synthesize negative lessons such as \texttt{avoid-wrong-focus}. This gives the agent a compact rule about what not to do, rather than only a recipe for what to do after the correct target has already been found.

\paragraph{Summary.}
On ScienceWorld \texttt{find-entity}, Auto-Dreamer reaches $72.5\%$ success with 7.4 steps per successful episode, while AWM reaches $26.2\%$ success with 20.0 steps per successful episode. The two methods end the run with nearly identical active-bank footprints: 13 entries and 795 tokens for Auto-Dreamer versus 10 entries and 816 tokens for AWM. This case shows that Auto-Dreamer's gains are not merely a consequence of compactness; they come from consolidating the right abstractions, including negative lessons from failure trajectories that success-only workflow induction fails to capture.